%% file: Main.tex
\documentclass{article}

\usepackage{arxiv}

\usepackage{listings}

\usepackage[utf8]{inputenc} 
\usepackage[T1]{fontenc}    
\usepackage{hyperref}       
\usepackage{url}            
\usepackage{booktabs}       
\usepackage{amsfonts}       
\usepackage{nicefrac}       
\usepackage{microtype}      
\usepackage{lipsum}		
\usepackage{graphicx}
\usepackage{natbib}
\usepackage{doi}
\usepackage{color}
\usepackage{amsmath}
\usepackage{subcaption}

\title{Dual Engines of Thoughts: A Depth-Breadth Integration Framework for Open-ended Analysis}

\author{ {\hspace{1mm}Fei-Hsuan Yu}\\
        NeuroWatt\\
	Intelligent Technology Research and Development \\
	\texttt{fiona@neurowatt.ai} \\
	\And
	{\hspace{1mm}Yun-Cheng Chou} \\
        NeuroWatt\\
	Intelligent Technology Research and Development \\
	\texttt{oro@neurowatt.ai} \\
    \And
	{\hspace{1mm}Dr. Teng-Ruei Chen} \\
        NeuroWatt\\
	Intelligent Technology Research and Development \\
	\texttt{luka@neurowatt.ai} \\
}

\renewcommand{\shorttitle}{NeuroWatt}

\hypersetup{
pdftitle={A template for the arxiv style},
pdfsubject={q-bio.NC, q-bio.QM},
pdfauthor={David S.~Hippocampus, Elias D.~Striatum},
pdfkeywords={First keyword, Second keyword, More},
}

\begin{document}
\maketitle

\begin{abstract}
We propose the Dual Engines of Thoughts (DEoT), an analytical framework for comprehensive open-ended reasoning. While traditional reasoning frameworks primarily focus on finding “the best answer” or “the correct answer” for single-answer problems, DEoT is specifically designed for “open-ended questions,” enabling both broader and deeper analytical exploration. The framework centers on three key components: a \emph{Base Prompter} for refining user queries, a \emph{Solver Agent} that orchestrates task decomposition, execution, and validation, and a \emph{Dual-Engine System} consisting of a Breadth Engine (to explore diverse impact factors) and a Depth Engine (to perform deep investigations). This integrated design allows DEoT to balance wide-ranging coverage with in-depth analysis, and it is highly customizable, enabling users to adjust analytical parameters and tool configurations based on specific requirements. Experimental results show that DEoT excels in addressing complex, multi-faceted questions, achieving a total win rate of 77--86\% compared to existing reasoning models, thus highlighting its effectiveness in real-world applications.
\end{abstract}

\keywords{Dual Engines of Thoughts \and Open-ended Analysis \and Multi-Agent Framework \and Reasoning Framework}

\section{Introduction}
In today's interconnected world, analyzing the implications of complex events requires a nuanced grasp of interconnected influences and their ripple effects. This is particularly evident in financial markets, where experienced analysts must evaluate how events influence market dynamics. Such analyses often require consideration of multiple dimensions, such as policy implications, technological advancements, and economic repercussions. Although large language models (LLMs) have demonstrated remarkable capabilities across various tasks, they face significant challenges when addressing open-ended questions that demand comprehensive analysis. These challenges include the need for multi-dimensional understanding, difficulty in capturing complex relationships between different aspects, and the balance between breadth and depth of analysis. 

Traditional approaches, whether human or automated, often struggle to provide both comprehensive coverage and detailed insights simultaneously. Automated systems tend to generate superficial analysis, while human analysts, although capable of deep analysis, are limited in their ability to explore multiple dimensions due to time and cognitive constraints.

Current reasoning frameworks such as Chain-of-Thought (CoT) \citep{wei2023}, Tree-of-Thoughts (ToT) \citep{press2023b}, and Graph-of-Thoughts (GoT) \citep{Besta_2024} have mainly focused on single-answer tasks or structured problem-solving. However, these frameworks lack the capacity to address multi-dimensional and open-ended questions effectively. A more advanced framework is necessary to bridge this gap and enable a holistic understanding of complex, open-ended scenarios. Similar to how human experts construct \textbf{brainstorming maps} to explore interconnected ideas dynamically, an effective system must be capable of balancing broad exploration with deep analysis in a structured yet flexible manner.

\subsection{Comparison with Traditional Frameworks}
To address these limitations, we propose \textbf{Dual Engines of Thoughts (DEoT)}, a framework that leverages two specialized engines for comprehensive analysis. The Breadth Engine explores diverse aspects of the analysis, while the Depth Engine conducts deep-dive investigations into specific areas of interest. This Dual-Engine approach enables both comprehensive coverage and in-depth understanding of complex scenarios. Table \ref{tab:framework-comparison} illustrates the key differences between DEoT and traditional frameworks:

\begin{table}[h]
\centering
\caption{Comparison of Traditional Frameworks and Dual Engines of Thoughts}
\label{tab:framework-comparison}
\begin{tabular}{p{2.5cm}||p{5cm}|p{5cm}}
\hline
\textbf{Aspect} & \textbf{Traditional Frameworks \newline (CoT/ToT/GoT)} & \textbf{Dual Engines of Thoughts \newline (DEoT)} \\
\hline
Objective & Step-by-step reasoning, single-task oriented & Simultaneous broad exploration and deep investigation \\
\hline
Processing \newline Structure & Single reasoning path or tree structure & Dual-engine architecture with coordinated analysis, similar to how humans construct \textbf{brainstorming maps} to dynamically explore ideas across different levels of abstraction. \\
\hline
Analysis Depth & Uniform depth across analysis & Adaptive depth based on significance and complexity \\
\hline
Scalability & Limited by structure complexity & Highly scalable through coordinated engine operation \\
\hline
Application Focus & Problem-solving tasks & Complex, multi-dimensional analysis scenarios \\
\hline
\end{tabular}
\end{table}

\subsection{Key Contributions}
The Dual Engines of Thoughts (DEoT) framework introduces a novel approach to open-ended reasoning by integrating both breadth and depth analysis. Our key contributions include:

\begin{itemize}
    \item \textbf{Dual-Engine Architecture:} DEoT features a Breadth Engine for multi-dimensional exploration and a Depth Engine for in-depth investigations. An Engine Controller dynamically balances both engines, ensuring comprehensive yet focused analysis.

    \item \textbf{Modular Analytical Framework:} DEoT’s flexible Base Prompter, Solver Agent, and Analysis Toolbox allow efficient query structuring, task execution, and tool-based reasoning. The system supports customizable tools, such as news search, event extraction, historical analysis, and direct reasoning, improving adaptability.

    \item \textbf{News-to-Question (N2Q) Dataset:} We constructed a benchmark dataset named N2Q, which contains follow-up questions derived from multi-domain news articles. This dataset simulates real-world open-ended analytical scenarios and serves as a valuable resource for future research.

    \item \textbf{Multi-Criteria Evaluation System:} We introduce an evaluation system using GPT-4o as a judge, assessing analytical performance based on five key criteria. A Dual Comparison Test ensures fairness and reproducibility, setting a new standard for open-ended reasoning evaluation.

     \item \textbf{Analytical Performance:}
    DEoT outperforms GPT-4o and Perplexity AI (Llama-3.1-sonar-small-128k-online), demonstrating:
    \begin{itemize}
        \item 77–86\% for total win rate, showing overall superiority in analytical tasks.
        \item 85–93\% in analytical depth, excelling in multi-layered reasoning.
        \item 70–75\%  logical coherence, ensuring structured and consistent outputs.
        \item 84–92\% in innovation, fostering creative insights through its dual-engine mechanism.
        \item Strong scalability, making it suitable for Biomedicine, Geopolitics, Economics, and Industry.
    \end{itemize}

\end{itemize}

This work establishes a robust foundation for advancing open-ended reasoning frameworks and highlights the potential of dual-engine architectures for multi-faceted analytical tasks.

\subsection{Framework Application}
While initially motivated by the challenges in financial market analysis, the DEoT framework demonstrates remarkable versatility in various domains that require comprehensive analysis. The system's ability to process complex events and generate multi-dimensional insights makes it a valuable tool for a wide range of applications, including investment impact analysis, technology trend monitoring, social impact assessment, scientific research implications, and policy evaluation.

\section{Related Works}

\subsection{Chain-of-Thought and Multi-Step Reasoning}

Multi-step reasoning is essential for LLMs in solving complex tasks. Chain-of-Thought (CoT) first demonstrated LLMs' reasoning potential by breaking problems into intermediate steps \citep{wei2023}. Zero-Shot CoT reduced the need for examples by using simple prompts such as "Let’s think step by step" \citep{kojima2023}. To improve result stability, Self-Consistency aggregates multiple reasoning paths \citep{wang2023}, while Least-to-Most Prompting guides models to progressively solve subtasks \citep{zhou2023}.

Beyond single-path methods, Tree-of-Thought (ToT) explores multiple reasoning paths with a tree structure \citep{long2023}, and Graph-of-Thought (GoT) extends this to nonlinear, graph-based reasoning for greater flexibility \citep{Besta_2024}. Despite these advancements, single-path or static logic remains limited for large-scale tasks, highlighting the need for multi-agent systems and tool integration to further enhance reasoning capabilities.

\subsection{Multi-Agent Systems}

Multi-Agent Systems (MAS) enable collaboration among multiple autonomous agents to solve complex tasks efficiently. AutoGen integrates LLMs, tools, and human inputs to facilitate dynamic task coordination and execution through agent conversations \citep{wu2023b}. AutoAgents further automate agent generation for reasoning and collaborative problem-solving \citep{chen2024}.

To enhance real-world reasoning, Sibyl introduces a lightweight agent framework that employs collaborative debate and global workspaces to tackle complex benchmarks \citep{wang2024b}. In addition, hierarchical approaches such as Agent-Oriented Planning \citep{li2024} and Hierarchical Multi-Agent Workflows \citep{liu2024}, focus on task decomposition and workflow optimization to improve coordination efficiency.

Recent surveys highlight the planning capabilities of LLM agents, emphasizing task adaptability and dynamic decision-making as key challenges for MAS development \citep{huang2024}. These advancements provide a foundation for DEoT, which integrates MAS principles with tool-driven reasoning for flexible, multi-dimensional analysis.

\subsection{Tool-Enhanced Reasoning Frameworks}

Tool integration enables LLMs to enhance reasoning and execution by interacting with external tools. ReAct first proposed combining reasoning with tool usage, allowing LLMs to dynamically plan and act based on intermediate feedback \citep{yao2023}. Similarly, HuggingGPT positioned LLMs as task orchestrators, integrating tools within the Hugging Face ecosystem to solve complex AI workflows \citep{NEURIPS2023}.

To improve tool retrieval efficiency, methods like Completeness-Oriented Tool Retrieval focus on selecting optimal tools for task execution \citep{Qu_2024}, while EASYTOOL simplifies tool usage through concise instructions to minimize ambiguity \citep{yuan2024}. Frameworks such as ToolPlanner introduce path planning and feedback mechanisms to refine multistep tool interactions \citep{wu2024b}, whereas ControlLLM leverages graph-based search for dynamic tool augmentation \citep{liu2023}.

Recent surveys \citep{shen2024, masterman2024} provide a comprehensive overview of LLMs with tools, highlighting challenges such as tool selection, dynamic feedback integration, and scalability. These studies form the basis for advanced frameworks like DEoT, which leverage tool integration to enable flexible, multi-dimensional analysis.

\section{Methodology}

\subsection{DEoT Framework Overview}
The DEoT framework introduces a novel approach to open-ended analysis by combining broad exploration with focused investigation. Traditional analysis methods often struggle to balance comprehensive coverage and deep understanding, either providing surface-level insights across many aspects or in-depth analysis with a limited scope. DEoT addresses this challenge through its innovative dual-engine architecture, designed to dynamically balance between breadth and depth of analysis.

At its core, DEoT relies on a \textbf{Base Prompter} to clarify user queries and employs a \textbf{Solver Agent} for processing complex analytical tasks, coordinated by an \textbf{Engine Controller} that determines the direction of the analysis. This design enables the system to adapt its analytical approach based on the current state and requirements of the analysis, similar to how human experts alternate between exploring broader implications and conducting detailed investigations of specific aspects, much like constructing a brainstorming map to visualize interconnected ideas. Through this adaptive approach, DEoT effectively handles complex analytical tasks while maintaining both breadth of coverage and depth of understanding. 

\subsection{Inspiration from Mind Mapping}
The conceptual inspiration for the Dual Engines of Thoughts (DEoT) framework partly arises from how humans often visualize problems usiing mind maps or brainstoriming diagrams. When people tackle a complex question, they commonly place the core issue in the center of a diagram, then radiate outward with multiple branches representing distinct perspectives or subquestions. This approach naturally balances breadth and depth: each branch expands a different facet of the problem (breadth), while crucial or intricate nodes warrant deeper exploration (depth).

In desinging DEoT, we aim to emulate this structured yet flexible process. The Breadth Engine systematically identifies and explores diverse impact factors-akin to drawing multiple branching paths in a mind map-while the Depth Engine zeroes in on specific nodes requiring a closer examination. The Engine Controller further embodies the metacoginitve decsion-making process, akin to how human analysts decide when to branch outward or probe more deeply into a particular train of thought.

By incorporating these mind-mapping principles, DEoT offers a framework that not only handles complex, multi-dimensional queries but also aligns with the intuitive ways in which humans brainstorm and integrate information. As a result, users can navigate intricate problem spaces with clarity—just as they might when sketching out a brainstorming map—while still benefiting from systematic computational strategies and tool integration.

\subsection{Framework Architecture}

\begin{figure}[h]
    \centering
    \includegraphics[width=0.98\textwidth]{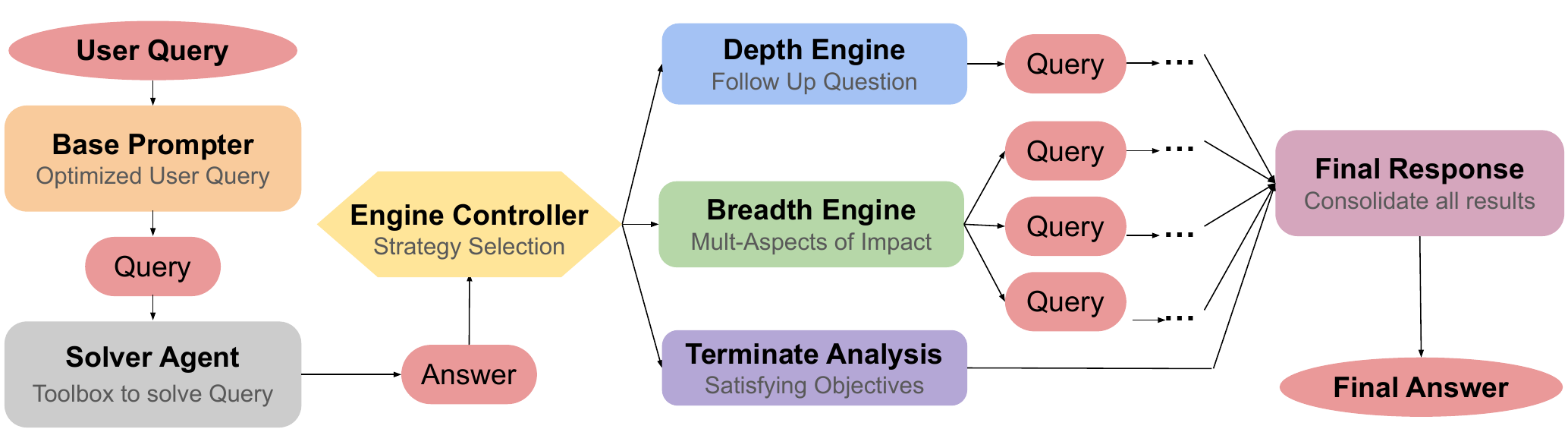}
    \caption{Dual Engines of Thoughts Analytical Framework}
    \label{fig:Analytical Framework}
\end{figure}

\subsubsection{Core Components}
The DEoT framework integrates three primary components: the Base Prompter, Solver Agent, and Dual-Engine System, each playing a crucial role in the analytical process.The analytical framework of DEoT is shown in Figure~\ref{fig:Analytical Framework}.

\paragraph{Base Prompter.}
The Base Prompter serves as the system’s entry point, transforming raw user input into well-structured analytical queries through comprehensive Query Optimization. It employs strategies such as refining vague queries, adding temporal context, clarifying ambiguous terms, and standardizing entity names. Additionally, it includes Error Handling mechanisms that analyze failed attempts and suggest corrections while preserving the core intent of the query. This ensures the subsequent analysis stages receive properly formatted, contextually enriched, and error-free input, thereby improving the efficiency and accuracy of the reasoning process. Inspired by Zhang et al.'s Ask-before-Plan approach \citep{zhang2024}, our system further highlights the crucial role of query refinement in optimizing analytical tasks, ensuring more structured and effective problem-solving.

\paragraph{Solver Agent.}
The Solver Agent functions as the central processing unit of the framework, orchestrating the entire analytical process through several specialized components:

\begin{itemize}

\item[(1)] The \textbf{Planner} performs two main tasks: Task Planning decomposes complex queries into one to three manageable tasks, considering dependencies and execution order, and assigns them to specialized agents. Plan Validation ensures these task plans are complete, non-redundant, and correctly formatted before execution.
 
\item[(2)] The \textbf{Toolbox} integrates various specialized analytical agents and offers customization options, allowing users to tailor the toolbox to meet specific task requirements. In our implementation, the following tools are combined to provide a comprehensive analytical framework. It includes the News Searcher, which retrieves and processes recent news articles; the Event Extractor, which extracts key information from events; the History Analyzer, which examines historical patterns and precedents; the Information Search, which gathers supplementary data for analysis; and the Direct Reasoning, which performs logical analysis and inference.

\item[(3)]  The \textbf{Executor} consists of two main tasks: Task Execution ensures tasks are executed in the correct sequence based on their dependencies. Result Validation verifies the factual accuracy of outputs, preventing the propagation of incorrect information in the analysis process.

\end{itemize}

\paragraph{Dual-Engine System.}
The Dual-Engine system drives the analytical process through three integrated components working in concert. The Engine Controller serves as the strategic center, evaluating the current state of analysis to determine the optimal analytical path. It considers factors such as content complexity, layer depth, and analysis coverage when deciding between breadth or depth exploration. Based on the controller's decision, either the Breadth Engine or Depth Engine is activated. The Breadth Engine identifies and explores different aspects of the analysis, systematically generating coverage. For each analysis node, it typically identifies three distinct aspects and their corresponding queries for investigation. The Depth Engine, conversely, generates targeted follow-up questions for specific areas requiring deeper investigation, enabling thorough exploration of significant or complex aspects of the analysis of the analysis.

\subsubsection{Analysis Flow}
The analytical process in DEoT follows a systematic yet flexible flow that ensures thorough coverage while maintaining analytical rigor.

\paragraph{Initial Processing.}
The process begins when a user input passes through the \textbf{Base Prompter}, which optimizes and structures the query by refining vague queries, adding temporal context, and clarifying ambiguous terms. This refined query then enters the \textbf{Solver Agent}, where the Planner decomposes it into one to three essential tasks. Each task plan undergoes plan validation for completeness and non-redundancy before the Executor initiates task execution using appropriate tools from the comprehensive toolbox. After the Executor Service completes all tasks, the Validation System conducts fact-checking to ensure the accuracy and reliability of the analysis.

\paragraph{Engine Selection.}
Following the initial analysis, the Engine Controller evaluates the current state to determine the next analytical direction. This decision considers multiple factors including content complexity, current layer depth, and the presence of specific points requiring investigation. If the Breadth Engine is selected, it will identify three distinct aspects of the topic and generates corresponding analytical queries. Conversely, if Depth Engine is chosen, it will generate one follow-up question targeting specific areas of interest. 

\paragraph{Iterative Processing.}
The generated queries, regardless of their originating engine, are reprocessed by the Solver Agent through iterative cycles. Each query-answer pair is treated as a node within the analytical process. The initial \textit{User-query} and its \textit{Answer} form the first layer of the analysis. Subsequent layers are constructed iteratively by the Dual-Engine System, which generates new questions based on insights from the previous layer. This layered approach facilitates a structured and dynamic exploration of the problem, allowing the analysis to evolve and refine as it progresses. This iterative cycle continues until one of the termination conditions is met: reaching the maximum layer depth, achieving the maximum node count, or satisfying all analysis objectives.

\begin{figure}[h]
    \centering
    \includegraphics[width=0.7\textwidth]{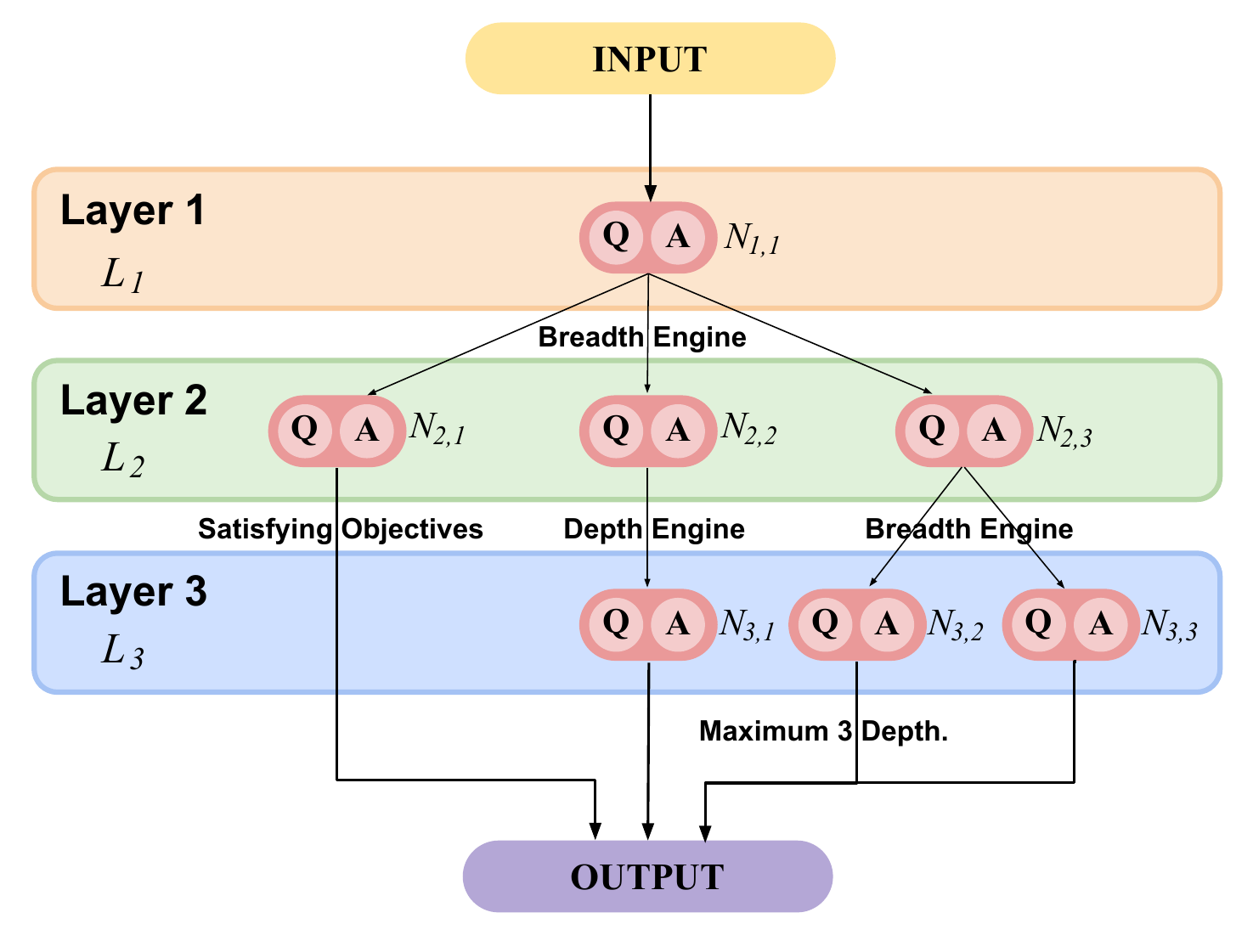}
    \caption{Dual Engines of Thoughts Tree Graph}
    \label{DEoT_Tree_Graph}
\end{figure}

Finally, the \textbf{Final Response Agent} consolidates information from all nodes to produce a comprehensive report that answers the user's original query in a structured and complete format. An example of the tree graph representing DEoT's nodes and layers is shown in Figure~\ref{DEoT_Tree_Graph}.

\subsection{Solver Agent Architecture}
The Solver Agent represents the core processing unit of the DEoT framework, designed to systematically break down complex analytical tasks into manageable components while ensuring rigorous quality control. This agent integrates Planner, Toolbox, Executor, and validation mechanisms.

\subsubsection{Planner}
The task Planner and Plan Validation process in the Planner follows a robust two-phase approach to ensure both completeness and accuracy. 

\paragraph{Task Planning.}
Building on research in MAS task decomposition, where agent-oriented planning enhances LLMs' problem-solving capabilities \citep{NEURIPS2023, li2024}, the Planner combines the user query with detailed descriptions of Toolbox tools to tailor sub-tasks effectively. It decomposes complex queries into 1-3 essential tasks, selects suitable analytical agents, and generates executable inputs. The planning process establishes clear task dependencies and execution order, ensuring a logical flow of analysis. Each generated sub-task includes a unique task ID, clear description, agent assignment, specific input parameters, and defined dependency relationships, enabling actionable and coherent outputs.

\paragraph{Plan Validation.}
Based on the key design principles for planning proposed by \cite{li2024}, we designed our Plan Validation process to ensure the accuracy of Task Planning. The Plan Validator employs a comprehensive three-tier validation system:

\begin{itemize}
    \item \textbf{Completeness Check}: Verifies that all key aspects of the query are covered, ensuring critical information is included and all tasks have proper agent assignments.
    \item \textbf{Non-Redundancy Check}: Evaluates tasks for overlapping requirements and unnecessary dependencies, optimizing the efficiency of the analytical process.
    \item \textbf{Format Correctness Check}: Ensures all task specifications meet required format standards.
\end{itemize}

When Plan Validation fails, the system initiates a plan regeneration process. This process incorporates specific feedback about the identified issues while preserving the original analytical objectives, ensuring that the revised plan addresses all validation concerns and remains aligned with the intended analysis goals.

\subsubsection{Executor}
The Executor combines Task Execution and Result Validation to ensure that analysis tasks are carried out systematically and produce accurate, reliable outputs.

\paragraph{Task Execution.}
The Task Execution manages the execution of tasks in a dependency-aware sequence. Each task is executed only when its dependencies are resolved, and the execution status (start, success, or failure) is logged at every step. Outputs from all tasks are integrated into a structured summary, providing a coherent basis for further analysis.

\paragraph{Result Validation.}
Result Validation focuses on verifying the factual accuracy and consistency of outputs. Building on the insights from \cite{lightman2023}, which highlights the critical role of sub-task result verification in enhancing overall analytical performance, this process involves:

    \begin{itemize} 
        \item \textbf{Task Validation}: Ensures factual accuracy of individual task outputs and records discrepancies with evidence.
        \item \textbf{Summary Validation}: Verifies combined results align with the original query and are free of inconsistencies.
    \end{itemize}

By identifying and addressing inaccuracies, the Executor ensures that all outputs are actionable, reliable, and aligned with the intended analysis goals.

\subsubsection{Analysis Toolbox}
The DEoT framework integrates a comprehensive set of analytical tools, each designed to handle specific aspects of the analysis process. These tools work in concert to ensure thorough and accurate analysis of complex scenarios.

\paragraph{News Search.}
The news search component, powered by Perplexity (Llama-3.1-sonar-small-128k-online), provides real-time access to current news and developments. It processes search queries with specific constraints on article count and relevance, ensuring focused and efficient information retrieval. This component plays a crucial role in gathering the most recent and relevant information for analysis. 

\paragraph{Event Extraction.}
The event extraction component, powered by the GPT-4o model, processes news content systematically, identifying and structuring key information. Leveraging GPT-4o’s advanced natural language understanding, it extracts essential details from raw text, including major events, involved entities, timestamps, locations, and causal relationships. 

\paragraph{History Analysis.}
The history analysis is also powered by GPT-4o model, it provides crucial contextual understanding by examining historical patterns and precedents. When analyzing current events, it identifies similar historical cases and examines their outcomes, enabling more informed analysis through pattern recognition and historical comparison. This historical perspective is particularly valuable in fields like financial analysis, where past events often provide crucial insights for understanding current situations.

\paragraph{Information Search.}
The information search component, powered by Perplexity AI (Llama-3.1-sonar-small-128k-online), serves as a supplementary knowledge acquisition system. When analysis reveals knowledge gaps or requires additional context, this component automatically formulates and executes targeted queries to gather relevant supplementary information. By seamlessly integrating new information into the analysis process, it helps maintain comprehensive and well-informed analysis.

\paragraph{Direct Reasoning.}
The direct reasoning, powered by the GPT-4o model, applies logical analysis and inference to synthesize insights from gathered information. It processes the outputs from other analytical tools to generate coherent analytical conclusions, identify potential implications, and suggest areas for deeper investigation. This component ensures that raw information is transformed into meaningful insights through systematic reasoning and analysis.

\subsection{Dual-Engine System}

The Dual-Engine System is the cornerstone of the DEoT framework, combining two complementary analytical engines—the Breadth Engine and the Depth Engine. These engines work together to balance comprehensive exploration with detailed investigation, creating a flexible and holistic approach to solving complex, open-ended problems.

\subsubsection{Engine Controller}

The Engine Controller is the decision-making core of the Dual-Engine System. It dynamically evaluates the context and determines the best course of action at each stage: whether to activate the Breadth Engine, Depth Engine, or terminate the analysis. Key factors guiding these decisions include:

\begin{itemize}
    \item \textbf{Content Complexity}: Assesses how diverse or intricate the problem is to determine whether broader exploration or deeper focus is needed.
    \item \textbf{Progress Metrics}: Monitors the current analytical layer relative to the maximum depth to ensure balanced and efficient progress.
    \item \textbf{Unresolved Points}: Identifies gaps in knowledge or unanswered questions that require further investigation.
    \item \textbf{Resource Utilization}: Evaluates the use of computational and time resources to prioritize high-value tasks and maintain efficiency.
\end{itemize}

\subsubsection{Breadth Engine}

The Breadth Engine is designed to explore different aspects of a problem systematically, making it particularly effective during the initial stages of analysis. This engine identifies key dimensions of the issue, such as economic, social, political, technological, or environmental factors, ensuring that all relevant perspectives are considered. It generates specific and actionable follow-up queries for each identified aspect, enabling a thorough exploration of the topic. Additionally, the Breadth Engine assigns priority levels to each aspect based on its importance and potential impact, ensuring that resources are allocated efficiently. By establishing a broad foundation, this engine provides a comprehensive understanding of the problem space.

\subsubsection{Depth Engine}

The Depth Engine focuses on detailed and targeted investigations of specific issues identified by the Breadth Engine or the Engine Controller. It is particularly effective in uncovering complex details and resolving intricate questions. This engine develops precise and thoughtful follow-up questions to address unresolved points or delve deeper into critical aspects. Through an iterative process, the Depth Engine refines its focus with each layer, generating increasingly detailed insights. It ensures that the questions are clear, actionable, and relevant, avoiding redundancy or overly broad scopes. By providing in-depth analysis, the Depth Engine ensures that the most important aspects of the problem are thoroughly explored and understood.

\subsubsection{System Integration and Termination}

The integration of the Breadth and Depth Engines is orchestrated by the Engine Controller, ensuring smooth transitions and collaboration between the two. The system operates iteratively, with outputs from one engine often informing the other. For example, the Breadth Engine may highlight key areas for investigation, which the Depth Engine then explores in greater detail.

In practical implementation, early stages of analysis prioritize the Breadth Engine to establish comprehensive coverage. In later stages, the Depth Engine is encouraged to refine and deepen the investigation.

The Engine Controller determines when to conclude the analysis by evaluating whether:

\begin{itemize}
    \item All key aspects and questions have been adequately addressed.
    \item The maximum allowable depth or breadth has been reached.
    \item Additional analysis would yield diminishing returns.
\end{itemize}

By dynamically balancing breadth and depth, the Dual-Engine System ensures the DEoT framework provides both broad and actionable insights, setting a high standard for addressing open-ended, multi-dimensional analytical challenges.

\begin{figure}[h]
    \centering
    \includegraphics[width=0.5\textwidth]{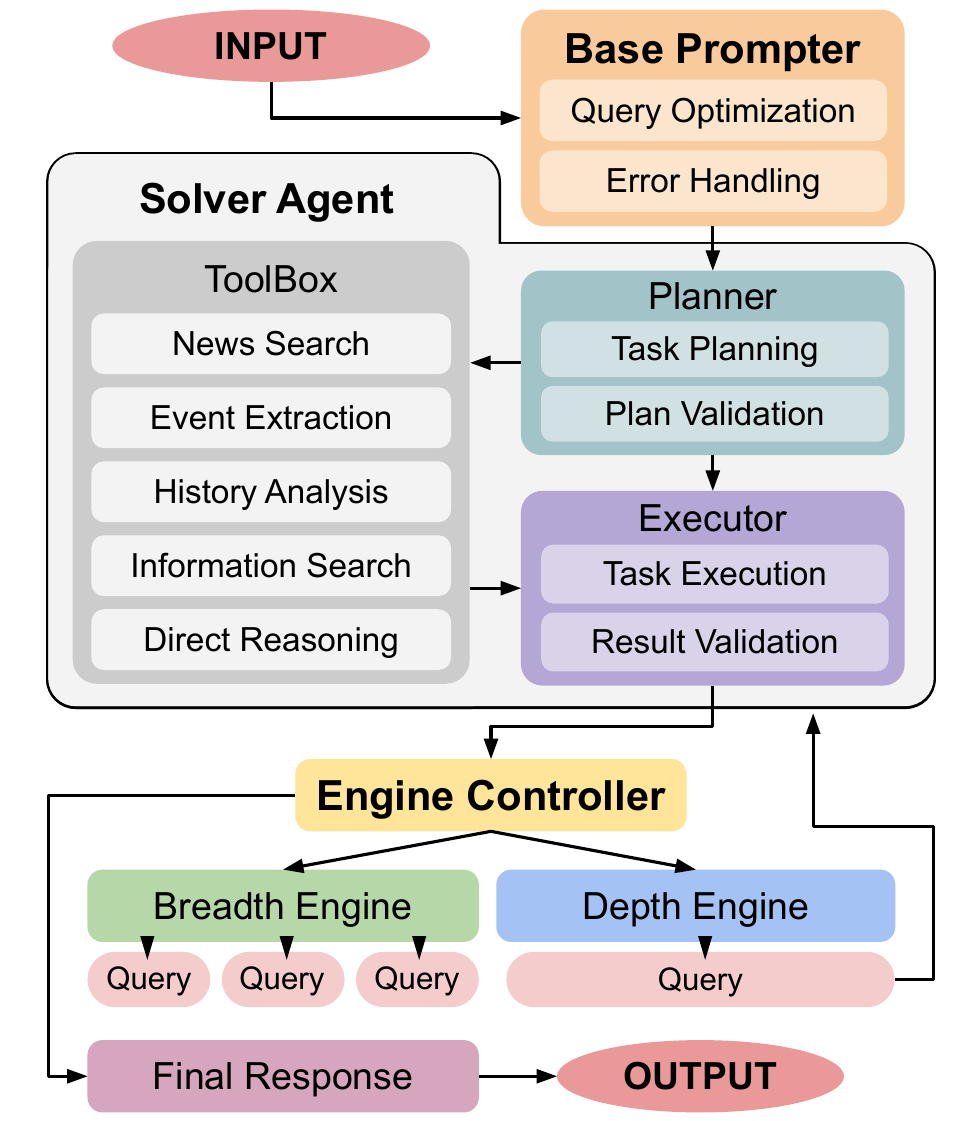}
    \caption{Dual Engines of Thoughts Analysis Flow}
    \label{DEoT_Analysis_Flow}
\end{figure}

In summary, the DEoT framework integrates multiple components, all orchestrated by the Engine Controller to ensure a seamless analytical workflow. The overall structure and interactions between these components are depicted in Figure~\ref{DEoT_Analysis_Flow}, which illustrates the complete analytical process from input to output.

\section{Experiments Design and Evaluation}
This study aims to comprehensively evaluate the performance of DEoT in open-ended question analysis and compare it with GPT-4o and Perplexity AI (Llama-3.1-sonar-small-128k-online). To ensure scientific rigor and fairness in evaluation, we constructed the News-to-Question (N2Q) dataset, which spans five major domains: Biomedicine, Economics, Geopolitics, Industry, and Technology, specifically designed to assess the performance of LLMs in multi-domain open-ended problem-solving.

For evaluation methodology, we adopt the LLM-as-a-judge approach proposed by Madaan et al. \cite{Neurips2023_2} and utilize GPT-4o as the core evaluation system. The models are systematically analyzed using five evaluation criteria: Analytical Depth, Specific Arguments, Innovation, Practicality, and Logical Coherence. Additionally, to enhance evaluation reliability, we introduce a Dual Comparison Test, which reduces potential biases and ensures result consistency and reproducibility.

\subsection{Experimental Objectives}

Despite recent advancements in LLMs, evaluating their ability to address open-ended problems remains a major challenge. Traditional single-answer tasks (e.g., question-answering systems) fail to fully assess LLMs’ potential in complex reasoning, causal analysis, and multi-layered decision-making. Therefore, the objectives of this experiment include:

\begin{itemize}
    \item Evaluating the performance of DEoT in open-ended questions, analyzing its capabilities in analytical depth, logical reasoning, and innovation, and exploring its advantages over existing LLMs. 
    \item Comparing the adaptability and stability of DEoT, GPT-4o, and Perplexity AI (Llama-3.1-sonar-small-128k-online) across multiple knowledge domains, analyzing their performance differences and strengths in various question types.  
    \item Exploring the feasibility and reliability of LLMs as evaluation agents, using a systematic scoring mechanism to ensure fairness and standardization in the assessment process while providing a practical benchmarking methodology.
\end{itemize}

\subsection{Dataset and Testing Domains}

\subsubsection{News-to-Question Dataset}  
Existing LLM benchmarks primarily focus on single-answer tasks, such as Fact Verification and Reading Comprehension, or on scenarios requiring sequential reasoning, such as Text-Based Games and Web Navigation. However, there is a lack of comprehensive test frameworks for evaluating open-ended questions, such as trend forecasting, causal analysis, and decision-making recommendations. Therefore, we constructed the N2Q dataset, which is based on real-time news events, designed to assess LLMs’ deep reasoning capabilities and provide a cross-domain open-ended evaluation standard.

\paragraph{Dataset Characteristics.}
The N2Q dataset has the following key attributes:
\begin{itemize}
    \item \textbf{Timeliness}: Questions are derived from the latest news events, ensuring relevance to current societal issues. 
    \item \textbf{Independence}: Each question includes a news summary and a follow-up question, allowing LLMs to conduct reasoning without referencing the original article. 
    \item \textbf{Diversity}: Through topic categorization and content filtering, the dataset ensures broad coverage across multiple domains while preventing excessive redundancy or similarity in questions.
    \item \textbf{Open-endedness}: The questions lack a definitive answer, involving historical comparisons, causal reasoning, trend forecasting, and strategic recommendations, which require high-level reasoning.
\end{itemize}

\paragraph{Dataset Construction.}

\begin{enumerate}
    \item News Extraction: We collected news articles from sources like Reuters, BBC, and Financial Times within the past month, ensuring content timeliness. LDA topic modeling and similarity filtering were applied to ensure diversity, with 100 news articles selected per domain.
    \item Question Generation: GPT-4o was used to generate an open-ended question for each news article using a structured prompt (detailed in Appendix A1):  
        \begin{itemize}
            \item News Event Overview: A concise summary of the news content to ensure question independence.
            \item Follow-up Question: A deep, open-ended question based on the news content to ensure open-endedness.  
        \end{itemize}
    \item Human Verification: Ensuring that questions are challenging and relevant, effectively testing LLMs’ analytical abilities.  
\end{enumerate}

\begin{figure}[h]
    \centering
    \includegraphics[width=1\textwidth]{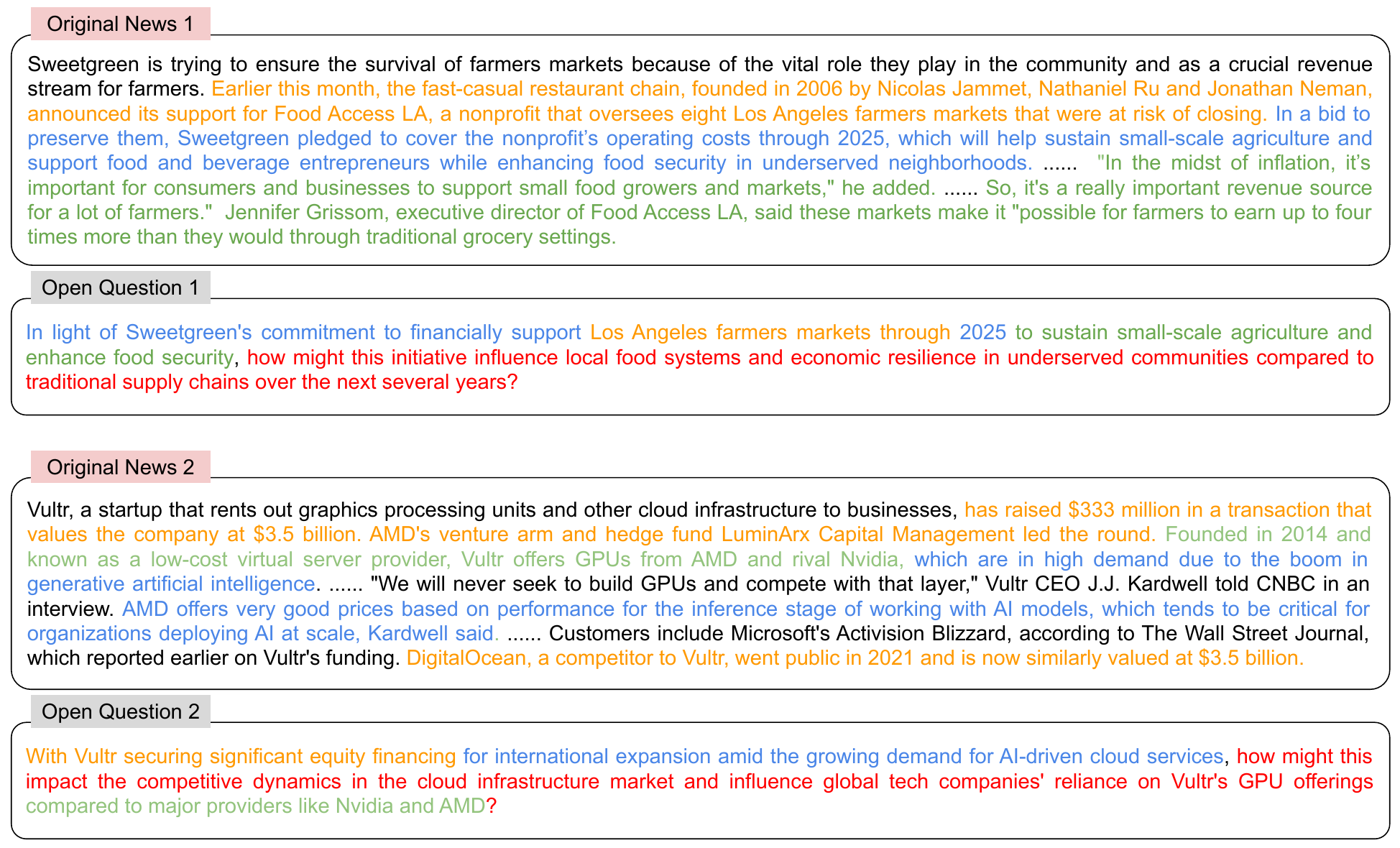}
    \caption{Two examples of open-ended question generation using N2Q on the news dataset. Colored texts indicate the structural composition of the questions: orange, green, and blue correspond to the Event Overview content derived from the news information, while red represents follow-up questions related to the event.}
    \label{fig:News-to-Question Dataset}
\end{figure}

Figure~\ref{fig:News-to-Question Dataset} illustrates two examples from the N2Q dataset, with different colored text corresponding to the news summary and follow-up question.

\subsubsection{Testing Domains} 

We selected five core domains to test LLMs' reasoning and adaptability in different scenarios:

\begin{itemize}
\item \textbf{Biomedicine}: News related to medical research, pharmaceutical developments, biotechnology, healthcare innovations, etc.
\item \textbf{Economics}: News related to finance, markets, economy, business, trade, etc.
\item \textbf{Geopolitics}: News related to international relations, political events, conflicts, policies that may affect markets, etc.
\item \textbf{Industry}: News related to manufacturing, energy, transportation, construction, and other industrial sectors.
\item \textbf{Technology}: News related to technological advancements, IT industry, software, hardware, AI, etc.
\end{itemize}

These domains were chosen due to their highly complex cross-domain causal relationships, requiring multi-layered reasoning capabilities from LLMs. In each domain, 100 questions were selected, totaling 500 open-ended questions, ensuring a sufficient dataset for comparative analysis of DEoT’s analytical performance across different disciplines.

The N2Q dataset provides a unique benchmark for evaluating LLMs’ cross-domain analytical capabilities through open-ended questions. Compared to existing datasets such as SQuAD and HotpotQA, N2Q emphasizes real-world news relevance, question diversity, and analytical depth.

\subsection{Model Comparison and Testing Procedure}

To evaluate the effectiveness of DEoT, we compare it with the following three systems:

\begin{itemize}
\item \textbf{DEoT} :  The proposed Dual-Engine reasoning system, which balances breadth and depth in analysis.
\item \textbf{GPT-4o}: One of the most advanced general-purpose LLMs, known for its strong natural language processing and logical reasoning capabilities.
\item \textbf{Perplexity AI (Llama-3.1-sonar-small-128k-online)} : An LLM integrating real-time retrieval capabilities, specializing in information synthesis and rapid response generation.
\end{itemize}

In the testing methodology, all models receive the same input questions with a uniform temperature setting of 0. The models are allowed to engage in open-ended reasoning without being constrained to a single-answer format. For DEoT, the maximum layer depth is set to 3, and the maximum node count is limited to 15. Finally, DEoT's performance is compared against baseline models using evaluation scores, and win rates are calculated across different evaluation criteria.

\subsection{Evaluation Design}
To ensure objectivity and standardization, we employ GPT-4o as the evaluation agent and assess the model outputs based on the five evaluation criteria. Recent studies indicate that advanced LLMs like GPT-4o align with human preferences in evaluation tasks by over 80\% \citep{zheng2023}, making it a reliable automated evaluation tool.

In many real-word scenarios, especially for open-ended tasks (e.g., essay writing, policy recommendations, or creative proposals), there is not single ``correct'' solution. Instead, human evaluators-like teachers grading essays-typically judge the output across multiple dimensions such as  grammar, coherence, persuasiveness, creativity, and factual accuracy. Inspired by this approach, we developed a multidimensional evaluation framework that captures various facets of the reasoning process. By accessing the models on several criteria, we can more holistically reflect their performance on open-ended analytic tasks, rather than relying on a binary correct-or-incorrect paradigm.

\subsubsection{Evaluation Criteria}
To ensure a comprehensive and fair evaluation, we compare DEoT with baseline models based on the following five core criteria:

\paragraph{Analytical Depth.}
Evaluates the model’s ability to deeply understand the underlying logic, implications, and relationships in a given problem space. Special attention is given to how the Depth Engine explores critical issues and integrates them into a coherent conclusion.

\paragraph{Specific Arguments.}
Assesses whether the model can effectively retrieve and summarize relevant data, examples, or historical analogies to substantiate its claims. The focus is on how well the Breadth Engine gathers supporting information and validates its accuracy and applicability.

\paragraph{Innovation.}
Measures the uniqueness and creativity of the model’s conclusions in real-world application scenarios, including policy recommendations, industry decision-making, and societal impact assessments. The goal is to avoid theoretical reasoning that lacks practical implementation value.

\paragraph{Practicality.}
Evaluates whether the proposed conclusions and solutions are applicable to real-world decision-making scenarios, considering constraints such as resources and contextual limitations rather than purely theoretical descriptions.

\paragraph{Logical Coherence.}
Examines the structural integrity and consistency of the reasoning process, ensuring arguments are logically sound and internally consistent. Additionally, it assesses how well the Breadth and Depth Engines integrate their findings into a unified analytical output rather than producing isolated responses.

Using these five criteria, the evaluation system systematically compares responses from different models to determine which one performs better in open-ended analytical reasoning.

\subsubsection{Evaluation System Architecture}
To minimize randomness and order bias in the evaluation process, we implement a Dual Comparison Test, which ensures fairness by rotating the input order for each model. The core mechanism operates as follows: In the first round, DEoT provides its answer first, followed by GPT-4o. In the second round, GPT-4o answers first, followed by DEoT.
The evaluation system follows these principles:
\begin{itemize}
\item Structured Evaluation: The evaluation system stores each model’s response in a structured format, ensuring transparency and traceability in the scoring process.
\item Multiple Scoring Mechanism: Each question undergoes two rounds of evaluation, where different models are assessed across the five evaluation criteria to ensure consistent and unbiased results.
\item Winner Selection: In each round, the system selects a winner per criterion, awarding 1 point to the superior model and 0 points to the other. The overall win rate serves as the key metric for measuring DEoT’s advantage over baseline models.
\end{itemize}
This methodology minimizes potential evaluation biases and provides a robust benchmark for DEoT’s performance in open-ended reasoning tasks.

This study uses the N2Q dataset to evaluate DEoT, GPT-4o, and Perplexity AI (Llama-3.1-sonar-small-128k-online) in five core domains. GPT-4o serves as the evaluation agent, applying five evaluation criteria and a Dual Comparison Test to enhance the fairness and stability of the assessment. These measures provide a precise benchmark for LLM capabilities in open-ended analysis. Ultimately, win rate analysis is used to determine DEoT’s advantages over the baseline models. In the next chapter (Chapter 5), we present detailed experimental results and conduct an in-depth discussion on DEoT’s performance in open-ended analytical tasks.

\section{Experimental Results}

To evaluate the effectiveness of the Dual Engines of Thoughts (DEoT) framework, we performed a comparative analysis against GPT-4o and Perplexity AI (Llama-3.1-sonar-small-128k-online). Our evaluation covered five core domains: Biomedicine, Economics, Geopolitics, Industry and Technology. The evaluation criteria included Total Win Rate, Analytical Depth, Specific Arguments, Innovation, Practicality, and Logical Coherence.

\subsection{Comparison with GPT-4o}

Table \ref{tab:gpt4o_comparison} presents the comparative analysis between DEoT and GPT-4o. DEoT achieved an overall win rate of 85. 5\% in all tested domains, with individual domain scores exceeding 81.5\%. Notably, DEoT achieved a high score in Geopolitics domain (89.0), indicating its advantage in international relations analysis, policy evaluation, and multi-level reasoning. This suggests that DEoT effectively integrates information and clarifies causal relationships.

\begin{table}[h]
\caption{Comparison between DEoT and GPT-4o.}
\centering
\begin{tabular}{|l|c|c|c|c|c|c|}
\hline
\textbf{Evaluation Criteria} & \textbf{Biomedicine} & \textbf{Economics} & \textbf{Geopolitics} & \textbf{Industry} & \textbf{Technology} & \textbf{Overall} \\
\hline
Total Win Rate & 86.5 & 87.0 & 89.0 & 83.5 & 81.5 & 85.5 \\
\hline
Analytical Depth & 92.5 & 94.0 & 93.5 & 93.5 & 89.0 & 92.5 \\
Innovation & 86.0 & 88.5 & 86.5 & 79.5 & 79.5 & 84.0 \\
Logical Coherence & 70.0 & 73.0 & 67.5 & 75.5 & 67.0 & 70.6 \\
Specific Arguments & 44.0 & 38.0 & 48.0 & 38.5 & 63.0 & 46.3 \\
Practicality & 23.5 & 18.0 & 22.0 & 23.5 & 27.5 & 22.9 \\
\hline
\end{tabular}
\label{tab:gpt4o_comparison}
\end{table}

\begin{figure}[h]
    \centering
    \begin{minipage}{0.48\textwidth}
        \centering
        \includegraphics[width=\textwidth]{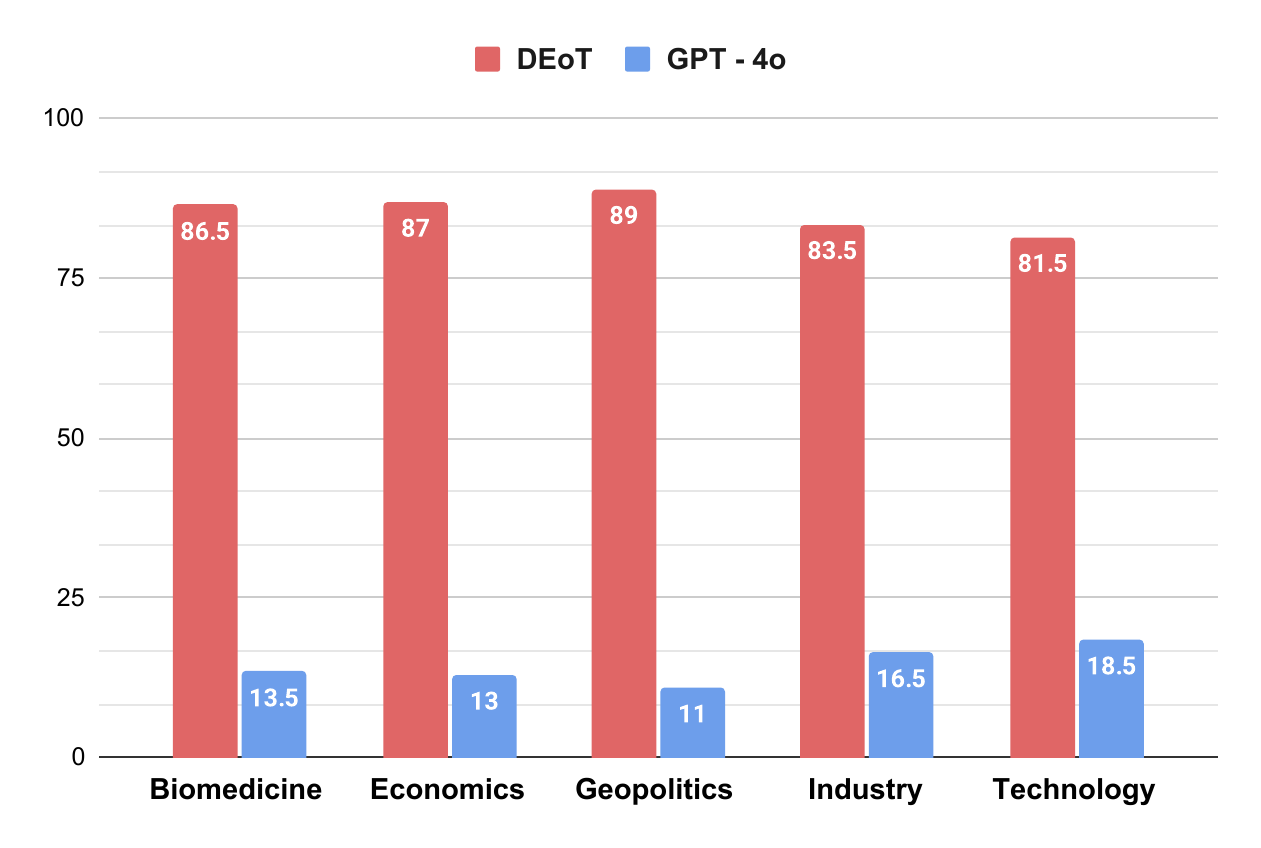}
        \caption{Comparison of DEoT and GPT-4o Based on Five Core Domains}
    \end{minipage}
    \hfill
    \begin{minipage}{0.48\textwidth}
        \centering
        \includegraphics[width=\textwidth]{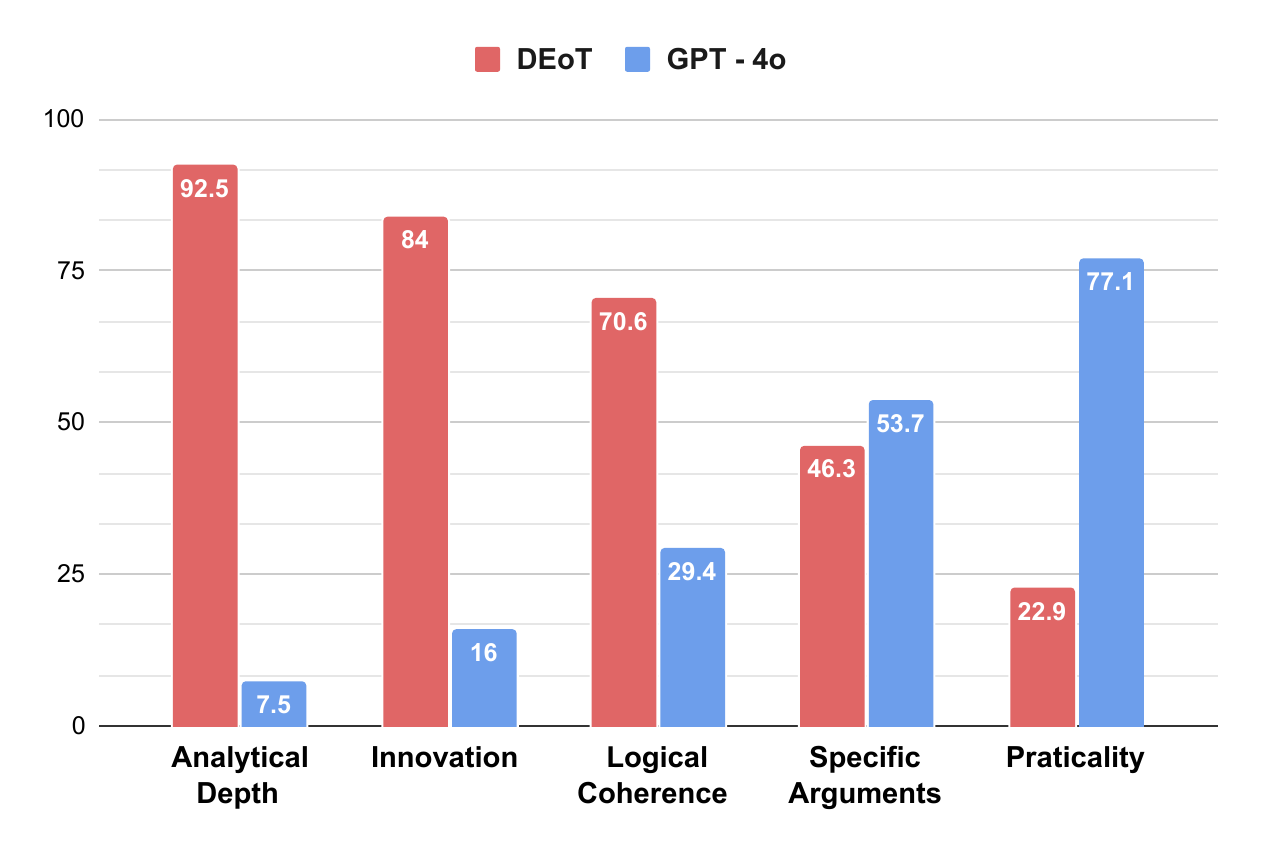}
        \caption{Comparison of DEoT and GPT-4o Based on Key Evaluation Criteria}
    \end{minipage}
    \label{fig:comparison}
\end{figure}

DEoT excelled in Analytical Depth (92.5), demonstrating its strong multilayered reasoning capability. Furthermore, DEoT outperformed GPT-4o in Logical Coherence (70.6) and Innovation (84.0), highlighting its competitive advantage in structured reasoning for complex issues. However, its score in Practicality (22.9) was relatively low, likely due to the depth of its reasoning, making some conclusions less immediately applicable to practical scenarios.

\subsection{Comparison with Perplexity AI (Llama-3.1-sonar-small-128k-online)}

Table \ref{tab:perplexity_comparison} presents the comparison between DEoT and Perplexity AI (Llama-3.1-sonar-small-128k-online). DEoT achieved an overall win rate of 77. 2\%, slightly less than GPT-4o. We attribute this to Perplexity AI’s real-time retrieval capabilities, which provide an advantage in the N2Q dataset. In Economics domain (83.0), DEoT demonstrated notable performance, suggesting its superiority in handling market dynamics, policy impacts, and complex economic modeling compared to Perplexity AI. This enables DEoT to analyze fluctuating economic environments more effectively and provide structured interpretations.

\begin{table}[h]
\caption{Comparison between DEoT and Perplexity AI (Llama-3.1-sonar-small-128k-online).}
\centering
\begin{tabular}{|l|c|c|c|c|c|c|}
\hline
\textbf{Evaluation Criteria} & \textbf{Biomedicine} & \textbf{Economics} & \textbf{Geopolitics} & \textbf{Industry} & \textbf{Technology} & \textbf{Overall} \\
\hline
Total Win Rate & 75.0 & 83.0 & 79.0 & 76.0 & 73.0 & 77.2 \\
\hline
Analytical Depth & 84.0 & 90.5 & 86.5 & 86.5 & 81.0 & 85.7 \\
Innovation & 90.0 & 94.5 & 92.5 & 91.5 & 91.5 & 91.9 \\
Logical Coherence & 72.0 & 78.5 & 76.0 & 75.0 & 71.0 & 74.5 \\
Specific Arguments & 3.5 & 5.5 & 2.0 & 4.5 & 2.0 & 3.5 \\
Practicality & 23.0 & 15.0 & 23.0 & 20.0 & 22.0 & 20.6 \\

\hline
\end{tabular}
\label{tab:perplexity_comparison}
\end{table}

\begin{figure}[h]
    \centering
    \begin{minipage}{0.48\textwidth}
        \centering
        \includegraphics[width=\textwidth]{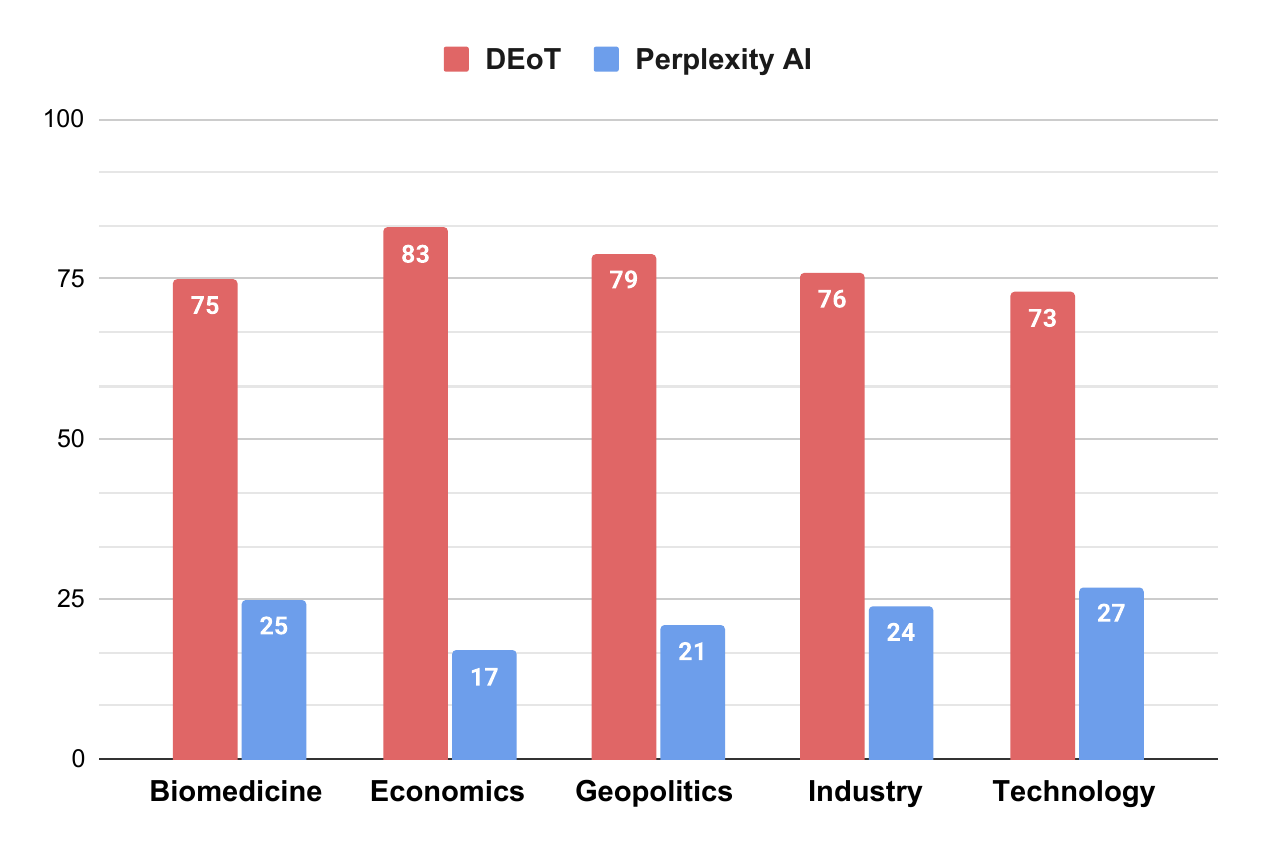}
        \caption{Comparison of DEoT and Perplexity Based on Five Core Domains}
    \end{minipage}
    \hfill
    \begin{minipage}{0.48\textwidth}
        \centering
        \includegraphics[width=\textwidth]{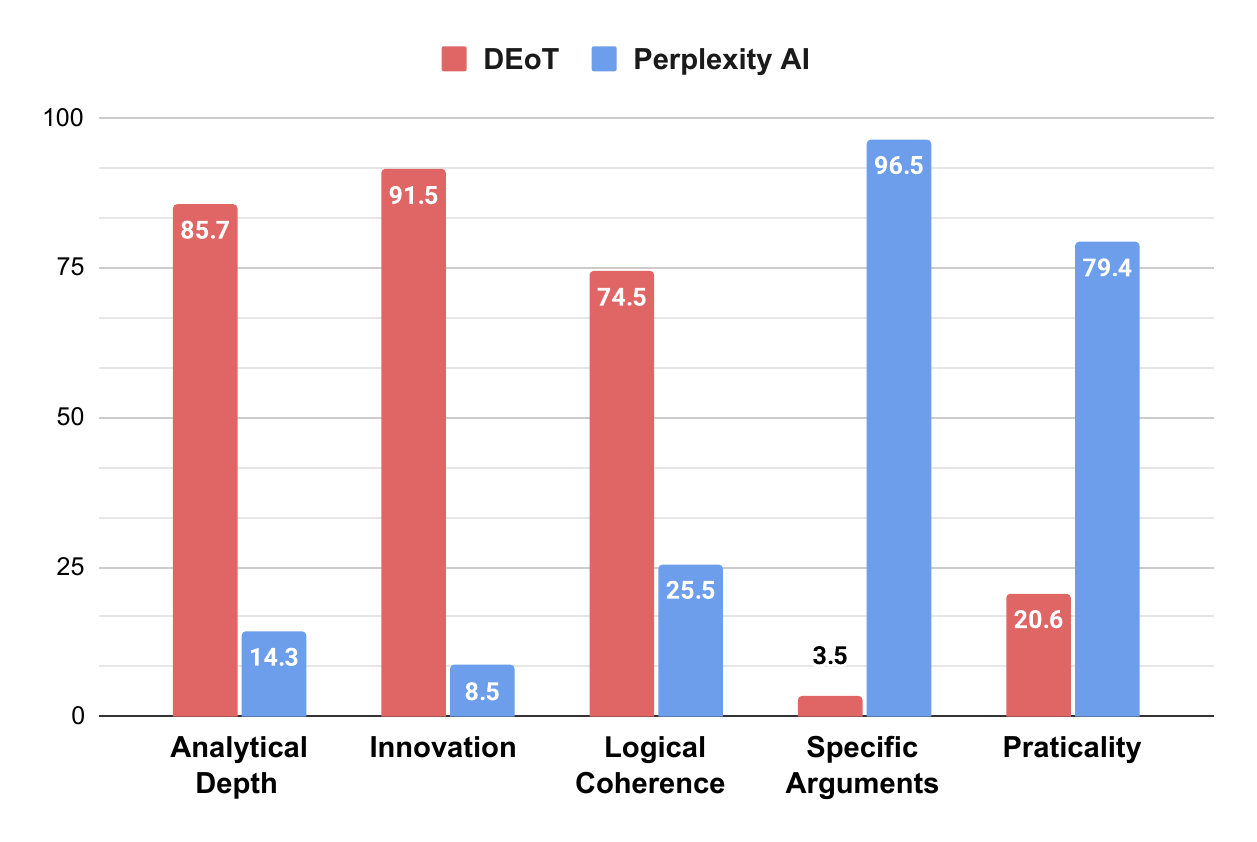}
        \caption{Comparison of DEoT and Perplexity AI (Llama-3.1-sonar-small-128k-online) Based on Key Evaluation Criteria}
    \end{minipage}
    \label{fig:comparison}
\end{figure}

DEoT excelled in Innovation (91.9), primarily due to its dual-engine architecture, which simultaneously explores breadth and depth, fostering more creative and diverse reasoning. This enables DEoT to generate novel insights and strategic recommendations more effectively than traditional single-path reasoning models. Additionally, DEoT surpassed Perplexity AI in Analytical Depth (85.7) and Logical Coherence (74.5), reinforcing its strength in multilevel reasoning and structured thinking. However, DEoT significantly underperformed in Specific Arguments (3.5), likely due to Perplexity AI’s reliance on real-time retrieval to enhance response accuracy and information completeness.

\subsection{Qualitative Comparison of Model Outputs}
\begin{figure}[htbp!]
    \centering
    \includegraphics[width=0.95\textwidth]{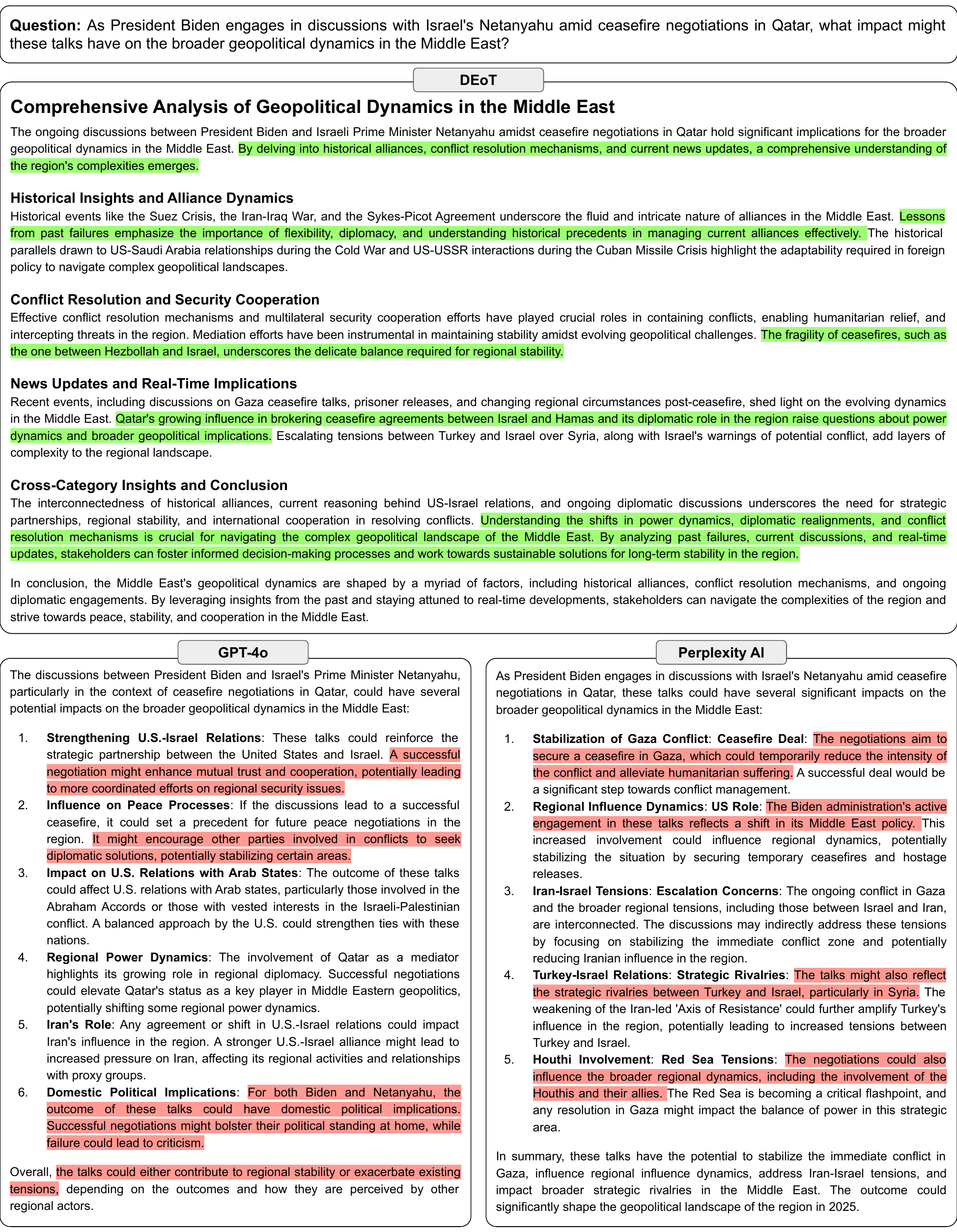}
    \caption{The figure compares DEoT, GPT-4o, and Perplexity AI in analyzing Biden and Netanyahu’s ceasefire talks, with green indicating strong responses and red weaker ones. DEoT’s dual-engine framework provides a more structured analysis by integrating historical context, conflict resolution, and real-time news, ensuring greater depth and breadth. This makes DEoT superior to models that rely solely on real-time data or single-layer reasoning.}
    \label{fig:DEoT_Comparison_Examples}
\end{figure}

While the quantitative results in the previous section offer a high-level performance overview, it is often insightful to examine specific examples of how different models respond to the same open-ended query. Figure~\ref{fig:DEoT_Comparison_Examples} presents a side-by-side comparison of responses from \textbf{DEoT}, \textbf{GPT-4o}, and \textbf{Perplexity AI: Llama-3.1-sonar-small-128k-online} to the question: 
\emph{``As President Biden engages in discussions with Israel's Netanyahu amid ceasefire negotiations in Qatar, what impact might these talks have on the broader geopolitical dynamics in the Middle East?''}

As highlighted by the color-coded annotations in Figure~\ref{fig:DEoT_Comparison_Examples}, \textbf{DEoT} delivers a more structured and multifaceted analysis, incorporating:
\begin{itemize}
    \item \textbf{Historical Context and Alliance Dynamics} -- referencing past conflicts (e.g., Suez Crisis, Iran-Iraq War) to illustrate the longstanding complexities of Middle Eastern alliances.
    \item \textbf{Conflict Resolution and Multilateral Cooperation} -- emphasizing the role of mediators, ceasefire fragility, and humanitarian relief efforts.
    \item \textbf{Real-Time Developments} -- integrating recent news updates (e.g., prisoner releases, the evolving role of Qatar) to show how current events might shift regional power balances.
    \item \textbf{Cross-Category Insights and Conclusions} -- weaving together policy, economic, and social dimensions into a cohesive outlook on regional stability.
\end{itemize}

By contrast, \textbf{GPT-4o} and \textbf{Perplexity AI} responses, while competent, tend to either focus on a narrower set of points or lack the integrated breadth-and-depth approach that DEoT demonstrates. GPT-4o provides multiple bullet points covering potential impacts, but not all are fully expanded upon or tied together into a larger narrative. Perplexity AI offers concise observations grounded in real-time data, yet does not delve as deeply into historical precedents or policymaking complexities. Overall, the dual-engine strategy of \textbf{DEoT} (Breadth + Depth) enables it to synthesize a more comprehensive understanding of geopolitical factors, reflecting its core design principle of balancing wide-ranging exploration with targeted deep-dive analysis.

\section{Discussion}

From a core domain perspective, DEoT excelled in the Economics and Geopolitics domains, likely due to the open-ended nature of these fields, where its dual-engine mechanism effectively supports multi-perspective analysis. However, the Technology domain demands deeper specialized expertise, placing a heavier emphasis on domain-specific knowledge and rapidly evolving paradigms—an aspect that remains challenging for DEoT’s generalized approach. In terms of evaluation criteria, DEoT exhibited notable strengths in Analytical Depth and Innovation, yet there is room for improvement in Practicality and Specific Arguments. Overall, the findings indicate that DEoT is particularly well-suited for tasks requiring in-depth reasoning and multi-layered exploration. Nevertheless, enhancing its capability to support specific arguments could further improve its applicability, ensuring that its insights provide greater practical and operational value.

One potential explanation for DEoT’s strong performance in open-ended analyses lies in how its Breadth and Depth Engines mirror human problem-solving—particularly brainstorming or mind-mapping approaches. By distributing cognitive load—first scanning a broad range of relevant factors (Breadth) before selectively drilling down (Depth)—DEoT effectively simulates the way people often tackle complex issues. In real-world applications such as multi-market investment decisions, policy assessment, and crisis management, this structured synergy allows domain experts to systematically explore risk factors, strategic trade-offs, and actionable outcomes. Nonetheless, DEoT’s lower scores in Practicality signal that the framework’s outputs may still require domain-specific curation or additional post-processing—for instance, specialized compliance checks, cost-feasibility estimators, or risk analysis modules—to translate theoretical insights into implementable solutions.

Moving forward, several opportunities exist to further refine DEoT. One avenue is to integrate specialized knowledge bases or domain-specific modules that can improve the system’s ability to retrieve and validate facts, thus bolstering its performance in fields like Technology or advanced biomedical research. Another is to develop adaptive heuristics for the Engine Controller, allowing it to dynamically balance breadth and depth based on problem complexity or user feedback. Additionally, exploring more advanced techniques for real-time data retrieval—especially in fast-evolving scenarios—could help address the system’s limitations in offering timely and relevant information. Finally, user-centric evaluations, where expert analysts interact with DEoT’s iterative outputs, may uncover new insights into how best to structure multi-layered analysis in genuine decision-making contexts, guiding further improvements in both system design and human–AI collaboration workflows.

\section{Conclusion}
In this paper, we introduced the Dual Engines of Thoughts (DEoT) framework, a novel approach for tackling open-ended analytical tasks through an integrated depth-breadth reasoning system. Our dual-engine architecture effectively balances broad exploration with focused investigation, ensuring comprehensive and structured reasoning across multi-dimensional problems.

Through extensive experiments using the News-to-Question (N2Q) dataset, we demonstrated DEoT’s superior analytical performance compared to state-of-the-art models like GPT-4o and Perplexity AI. DEoT excelled particularly in Analytical Depth 85-93\% and Innovation 84-92\%, highlighting its capability to generate insightful, multi-layered reasoning. Additionally, its structured approach led to notable improvements in Logical Coherence 70-75\%, ensuring a well-integrated analytical process. However, practical applicability 20-23\% and specific argument support 3-47\% remain areas for further enhancement.

The real-world applicability of DEoT extends across multiple domains, including finance, policy-making, social impact analysis, and technology forecasting. By leveraging its adaptive query structuring, tool integration, and multi-agent coordination, DEoT can provide more structured and actionable insights compared to traditional single-path reasoning frameworks.

Despite its strengths, DEoT still faces several limitations. Its reliance on pre-existing knowledge constrains specific argument retrieval, especially in specialized domains like biomedicine and semiconductor analysis. Moreover, while its depth-first approach enhances analytical richness, it may reduce practical applicability in real-world decision-making scenarios.

Future research will focus on enhancing real-time data retrieval capabilities, integrating domain-specific expert knowledge, and refining the Engine Controller to dynamically adjust reasoning depth based on context sensitivity. Additionally, expanding DEoT’s application to domains such as legal reasoning, medical diagnostics, and automated strategic analysis will further test its scalability and robustness.

In conclusion, DEoT represents a significant step forward in multi-dimensional, open-ended analysis. By systematically integrating depth and breadth reasoning, it establishes a new benchmark for comprehensive analytical frameworks, paving the way for more adaptive, structured, and insightful AI-driven reasoning systems.

\bibliographystyle{unsrtnat}
\bibliography{references}  

\newpage

\appendix

\input{appendix}

\end{document}

%% file: appendix.tex

\section{DEoT Case Studies}

\begin{figure}[h]
    \centering
    \includegraphics[width=1\textwidth]{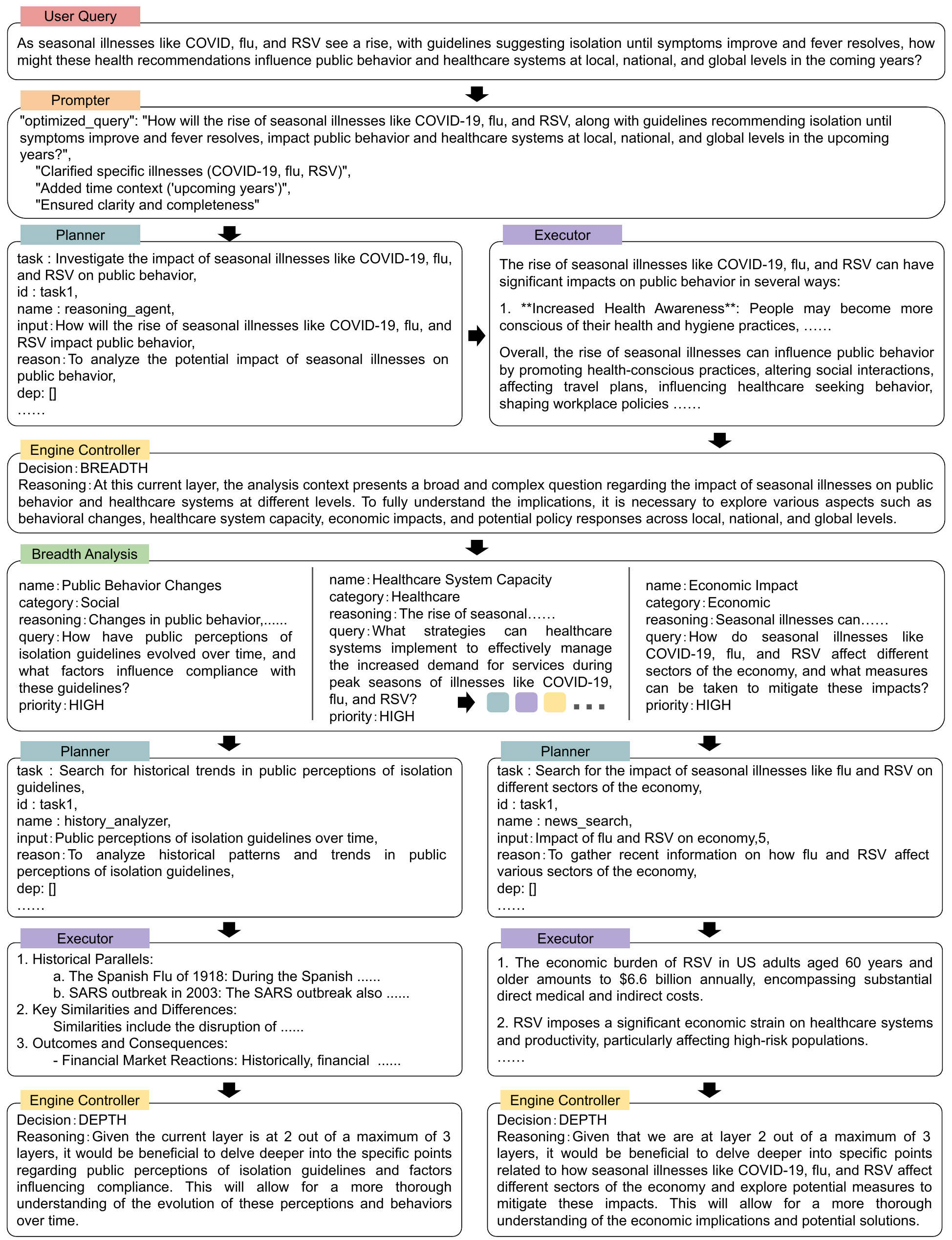}
    \caption{DEoT Case Studies Part-1}
    \label{fig:DEoT Case Studies Part-1}
\end{figure}

\begin{figure}[h]
    \centering
    \includegraphics[width=1\textwidth]{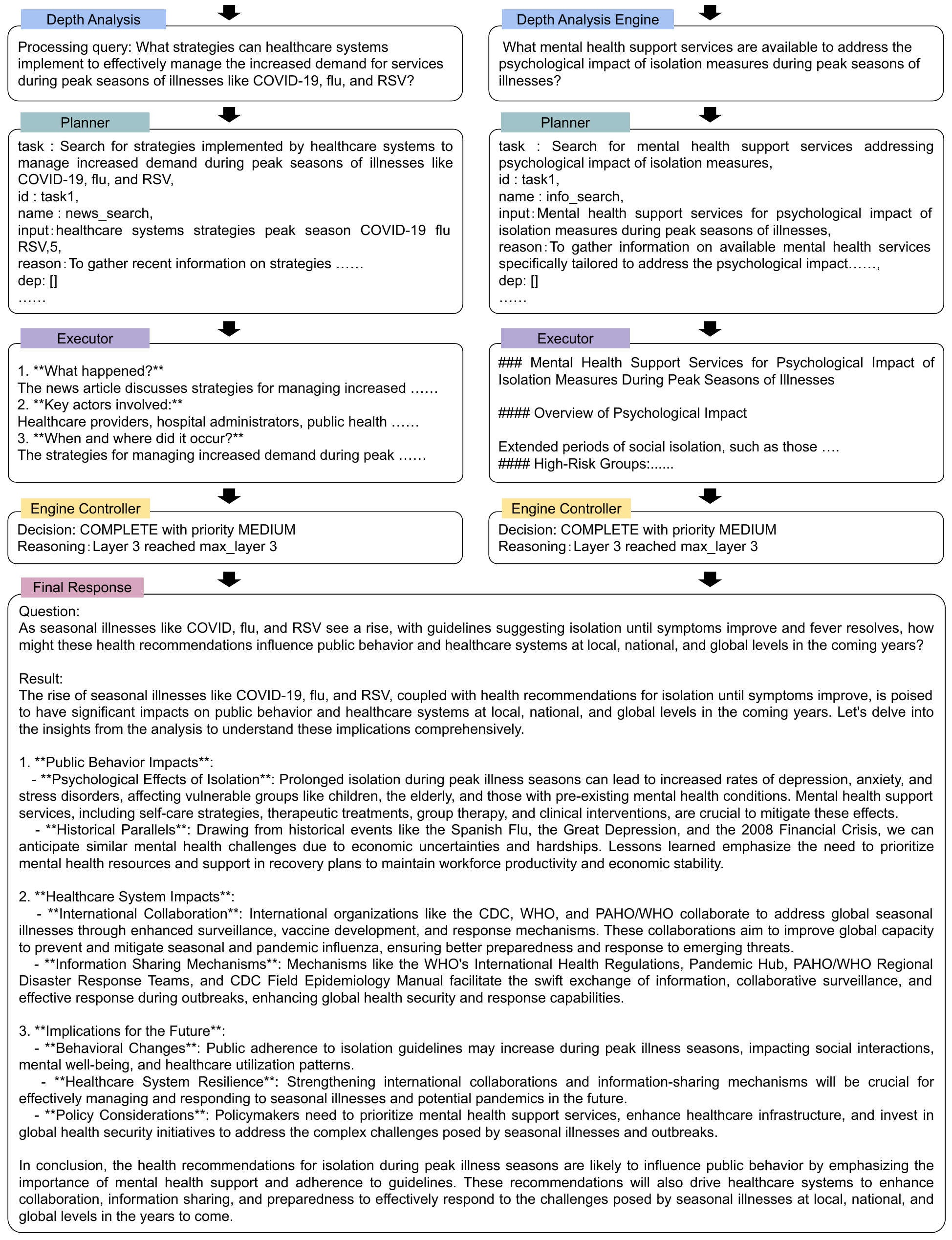}
    \caption{DEoT Case Studies Part-2}
    \label{fig:DEoT Case Studies Part-2}
\end{figure}

In this section, we show more case studies of DEoT for the N2Q datasets in Figure~\ref{fig:DEoT Case Studies Part-1} and Figure~\ref{fig:DEoT Case Studies Part-2}, respectively.

\section{Prompter Agent Prompts} The prompts used by the prompter agent, designed for input optimization and error handling, are provided for reference in Figure~\ref{Base prompter} and Figure~\ref{Error handling}.  These prompts help ensure that queries are clear, specific, and ready for execution.

\begin{figure}[ht]
\centering
\begin{lstlisting}[frame=single, basicstyle=\scriptsize\ttfamily, breaklines=true]
base_prompter:
 input_optimization:
   system: |
     You are an input optimization agent. Your task is to enhance the user's query.
     
     You must return a JSON object with EXACTLY these fields:
     {
       "optimized_query": "<enhanced version of the query>",
       "original_query": "<the original query>",
       "modifications": ["list of changes made"]
     }

     Optimization goals:
     1. Make vague queries more specific
     2. Add missing temporal context
     3. Clarify ambiguous terms
     4. Standardize entity names
     5. Ensure the query is clear and complete

     Example output:
     {
       "optimized_query": "What is Tesla (TSLA) stock's current performance in December 2024?",
       "original_query": "How's TSLA doing?",
       "modifications": [
         "Added full company name",
         "Added time context",
         "Specified stock performance metric"
       ]
     }

   user: |
     Query to optimize: {input}

\end{lstlisting}
\caption{Input Optimization Prompt for Base Prompter Agent}
\label{Base prompter}
\end{figure}


\begin{figure}[ht]
\centering
\begin{lstlisting}[frame=single, basicstyle=\scriptsize\ttfamily, breaklines=true]
 error_handling:
   system: |
     You are an input optimization agent handling errors. You need to:
     1. Understand the original optimization task
     2. Analyze the error and previous failed attempt
     3. Provide corrections while maintaining the query's core intent

     You must return a JSON object with EXACTLY these fields:
     {
       "optimized_query": "<corrected and enhanced version of the query>",
       "original_query": "<the original query>",
       "modifications": ["list of changes made"],
       "error_handling": {
         "original_error": "<the error message>",
         "correction_explanation": "<explanation of corrections made>",
         "previous_attempt_analysis": "<analysis of what went wrong in the previous attempt>"
       }
     }

     Example of successful error handling:
     {
       "optimized_query": "What is Tesla (TSLA) stock's current performance in December 2024?",
       "original_query": "How's TSLA doing?",
       "modifications": [
         "Added full company name",
         "Added time context",
         "Specified stock performance metric"
       ],
       "error_handling": {
         "original_error": "JSON parsing error in previous response",
         "correction_explanation": "Fixed JSON format and added missing required fields",
         "previous_attempt_analysis": "Previous attempt failed due to malformed JSON structure"
       }
     }

   user: |
     Original Query: {original_query}
     Error Encountered: {error_message}
     Previous Failed Result: {failed_result}

     Please analyze the error and previous attempt, then provide an optimized version of the query.
     Consider what went wrong in the previous attempt and ensure the response follows the correct format.
     Remember to maintain the core intent of the original query while fixing the issues.

\end{lstlisting}
\caption{Error Handling Prompt for Prompter Agent}
\label{Error handling}
\end{figure}

\section{Solver Agent Prompts}

The Solver Agent is composed of three main components: Planner, ToolBox, and Executor, each designed to collaboratively solve complex queries. The prompts used by these components are provided for reference in the corresponding figures.

\paragraph{Planner}
The prompts used by the planner agent, designed for task decomposition and plan validation, are provided for reference in Figure \ref{Task_Decomposition_1}, \ref{Task_Decomposition_2} and Figure \ref{Plan_Validator}. These prompts assist in breaking down complex queries into manageable tasks and validating the resulting plans for completeness and correctness.

\begin{figure}[ht]
\centering
\begin{lstlisting}[frame=single, basicstyle=\scriptsize\ttfamily, breaklines=true]
task_decomposition:
 meta_agent:
  system: |
    You are a task decomposition agent. Your responsibility is to break down the given query into subtasks and select the most appropriate agent for each task.

    Available agents:
    - news_search: Search and retrieve news articles and real-time information
      Input format: "query,number" (e.g. "Ukraine conflict,5")
    - event_extractor: Extract key events and their relationships from text
    - history_analyzer: Analyze historical patterns and similar cases
    - info_search: Search for supplementary information
    - llm: Generate answers using language model reasoning

    IMPORTANT CONSTRAINTS:
    1. Generate 1 ~ 3 tasks in total.
    2. Focus on the most essential tasks to answer the query
    3. For news_search tasks:
      - Input MUST be in format "query,number" (e.g. "Ukraine conflict,5")
      - Search for NO MORE THAN 5 articles (number must be 1-5)

    For each task, you need to:
    1. Select the most appropriate agent
    2. Generate specific input for that agent:
      - For news_search: "your query,number" (e.g. "Tesla stock price,3")
      - For other agents: normal text input
    3. Define dependencies if any

    Please output in the following JSON format:
    [{
      "task": "task_description",
      "id": "task_id",
      "name": "agent_name",
      "input": "specific input for the agent (for news_search use: query,number)",
      "reason": "detailed_reason_for_agent_selection",
      "dep": ["dependency_task_ids"]
    }]

    Example task using news_search:
    {
      "task": "Search for recent news about Tesla stock performance",
      "id": "task1",
      "name": "news_search",
      "input": "Tesla stock price performance,3",
      "reason": "To gather recent information about Tesla stock trends",
      "dep": []
    }

    Note:
    - Each task must have clear, executable input for the agent
    - Include all critical information (numbers, names, etc.)
    - Consider dependencies between tasks when sequencing tasks
    - For tasks that depend on previous results, specify how to use those results in the input field
    - Only output in JSON, without additional messages

  user: |
    Please decompose the following query into specific tasks with agent assignments and inputs:
    {input}

\end{lstlisting}
\caption{Task Decomposition Prompt for Planner Agent, Part 1.}
\label{Task_Decomposition_1}
\end{figure}

\begin{figure}[ht]
\centering
\begin{lstlisting}[frame=single, basicstyle=\scriptsize\ttfamily, breaklines=true]
 retry:
    system: |
     You are a task decomposition agent. Your responsibility is to break down the given query into subtasks and select the most appropriate agent for each task.

      Available agents:
      - news_search: Search and retrieve news articles and real-time information
        Input format: "query,number" (e.g. "Ukraine conflict,5")
      - event_extractor: Extract key events and their relationships from text
      - history_analyzer: Analyze historical patterns and similar cases
      - info_search: Search for supplementary information
      - llm: Generate answers using language model reasoning

      IMPORTANT CONSTRAINTS:
      1. Generate 1 ~ 3 tasks in total.
      2. Focus on the most essential tasks to answer the query
      3. For news_search tasks:
        - Input MUST be in format "query,number" (e.g. "Ukraine conflict,5")
        - Search for NO MORE THAN 5 articles (number must be 1-5)

      For each task, you need to:
      1. Select the most appropriate agent
      2. Generate specific input for that agent:
        - For news_search: "your query,number" (e.g. "Tesla stock price,3")
        - For other agents: normal text input
      3. Define dependencies if any

      Please output in the following JSON format:
      [{
        "task": "task_description",
        "id": "task_id",
        "name": "agent_name",
        "input": "specific input for the agent (for news_search use: query,number)",
        "reason": "detailed_reason_for_agent_selection",
        "dep": ["dependency_task_ids"]
      }]

      Example task using news_search:
      {
        "task": "Search for recent news about Tesla stock performance",
        "id": "task1",
        "name": "news_search",
        "input": "Tesla stock price performance,3",
        "reason": "To gather recent information about Tesla stock trends",
        "dep": []
      }

      Note:
      - Each task must have clear, executable input for the agent
      - Include all critical information (numbers, names, etc.)
      - Consider dependencies between tasks when sequencing tasks
      - For tasks that depend on previous results, specify how to use those results in the input field
      - Only output in JSON, without additional messages

    user: |
      Original Query: {original_query}
      Previous Feedback: {feedback}
      
      Please provide a revised task decomposition that:
      1. Addresses the feedback above
      2. Follows the required JSON format
      3. Ensures all tasks are properly linked and executable
      4. Maintains completeness and non-redundancy

\end{lstlisting}
\caption{Task Decomposition Prompt for Planner Agent, Part 2.}
\label{Task_Decomposition_2}
\end{figure}

\begin{figure}[ht]
\centering
\begin{lstlisting}[frame=single, basicstyle=\scriptsize\ttfamily, breaklines=true]
plan_validator:
   system: |
    You are a plan validator responsible for analyzing task plans for "completeness", "non-redundancy", and "format correctness".

    Your task is to perform three types of checks:

    1. Completeness Check:
    - Verify if the task set covers all key aspects of the initial query
    - Check if all critical info (numbers and names) is included
    - Examine if every task has at least one agent assigned
    - Ensure each task has clear, executable input
    - Verify if all necessary dependencies are properly set
    - Check if any key information is missing

    2. Non-redundancy Check:
    - Evaluate if any tasks contain overlapping information or requirements
    - Check if any tasks can be combined without losing clarity
    - Verify if dependencies are necessary and non-circular
    - Look for any duplicate work across tasks

    3. Format Correctness Check:
    - For news_search tasks:
      * Input MUST be in format "query,number" (e.g. "Brazil sugar exports,5")
      * Number of articles must be between 1 and 5
      * Query part cannot be empty
    - For all tasks:
      * Input field cannot be empty
      * Dependencies must reference existing task IDs

    If the plan needs no modifications, respond with:
    'The plan satisfies completeness and non-redundancy.'

    Otherwise, provide specific feedback including:
    - What aspect is problematic (completeness, redundancy, or format)
    - Which tasks are affected
    - Specific suggestions for improvement

   user: |
     Original Query: {original_query}
     Generated Task Plan: {task_plan}

     Please validate this task plan based on the criteria above.

\end{lstlisting}
\caption{Plan Validation Prompt for Planner Agent.}
\label{Plan_Validator}
\end{figure}

\paragraph{ToolBox}
The prompts used by the ToolBox components, including News Search (Figure \ref{News_Search}), Event Extractor (Figure \ref{Event_Extractor}), History Analyzer (Figure \ref{History_Analyzer}), Reasoning (Figure \ref{Reasoning}), and Info Search (Figure \ref{Info_Search}), are designed for tasks like information retrieval, event extraction, and logical reasoning. These prompts support efficient query resolution and analysis.

\begin{figure}[ht]
\centering
\begin{lstlisting}[frame=single, basicstyle=\scriptsize\ttfamily, breaklines=true]
news_search:
 search:
   system: |
     You are a news search agent for Bloomberg, CNBC and CBS News.
     Return articles in this format:
     ### [Title]
     **Source:** [Bloomberg/CNBC/CBS]
     **URL:** [url]
     **Summary:**
     [Brief factual summary]

   user: |
     Please find {needed_count} recent news articles about: {query}
     Requirements:
     1. Return exactly {needed_count} articles
     2. Only return articles from Bloomberg, CNBC or CBS
     3. Focus on articles directly related to the query
     4. Present in specified format
\end{lstlisting}
\caption{Prompt for News Search in ToolBox}
\label{News_Search}
\end{figure}

\begin{figure}[ht]
\centering
\begin{lstlisting}[frame=single, basicstyle=\scriptsize\ttfamily, breaklines=true]
event_extractor:
  extract:
    system: |
      You are an expert in extracting key events from news articles. Your task is to identify and summarize the 
      most important events, focusing on:
      1. What happened?
      2. Who are the key actors involved?
      3. When and where did it occur?
      4. What are the immediate consequences or reactions?
      5. Any numerical data or statistics mentioned?

      Present the extracted information in a clear, concise format. Prioritize factual information over speculation or opinion.
      Please reply in at most 300 words.

    user: |
      Extract key events from the following news article:

      {text}

\end{lstlisting}
\caption{Prompt for Event Extraction in ToolBox.}
\label{Event_Extractor}
\end{figure}

\begin{figure}[ht]
\centering
\begin{lstlisting}[frame=single, basicstyle=\scriptsize\ttfamily, breaklines=true]
history_analyzer:
  analyze:
    system: |
      You are a historical analysis expert focusing on financial, political, and business impacts 
      of events.
      Analyze events by comparing them with historical parallels and providing insights about:
      1. Financial market reactions
      2. Political ramifications
      3. Business sector impacts
      4. Investment strategies and policy responses

      Your analysis should be detailed, specific, and focused on practical implications.
      Keep responses under 300 words.

    user: |
      Analyze this event and provide historical parallels:
      Event: {event}

      Please cover:
      1. 2-3 similar historical events, emphasizing financial, political, and business relevance
      2. Key similarities and differences in economic and market impacts
      3. Outcomes and consequences:
         - Financial market reactions
         - Political ramifications
         - Business sector impacts
      4. Lessons and insights for current situation

\end{lstlisting}
\caption{Prompt for Historical Analysis in ToolBox.}
\label{History_Analyzer}
\end{figure}

\begin{figure}[ht]
\centering
\begin{lstlisting}[frame=single, basicstyle=\scriptsize\ttfamily, breaklines=true]

Reasoning:
  reason:
    system: |
      You are an analytical reasoning assistant. Provide clear, logical, and well-structured responses.
      Focus on:
      1. Direct answers to queries
      2. Clear explanations
      3. Logical reasoning
      4. Relevant examples when appropriate

      Keep responses clear and concise.

    user: |
      Please provide analysis and insights on:
      {query}

\end{lstlisting}
\caption{Prompt for Logical Reasoning in ToolBox.}
\label{Reasoning}
\end{figure}

\begin{figure}[ht]
\centering
\begin{lstlisting}[frame=single, basicstyle=\scriptsize\ttfamily, breaklines=true]
info_search:
  search:
    system: |
      You are an information search agent. Search and provide comprehensive information with sources.
      For each piece of information, you must include:
      1. The source (website, publication, or organization)
      2. Key information from that source
      3. Any relevant data or statistics

      Format requirements:
      - Structure the response with clear sections
      - Clearly indicate the source for each piece of information
      - Prioritize reliable and authoritative sources

    user: |
      Please search and provide detailed information about:
      {query}

      Remember to:
      1. Include sources for all information
      2. Focus on factual data and verified information
      3. Present information in a clear, organized format
      4. Response in 200 words

\end{lstlisting}
\caption{Prompt for Information Search in ToolBox.}
\label{Info_Search}
\end{figure}

\paragraph{Executor}
The prompts used by the Executor components, designed for task execution and result validation, are provided for reference in Figure \ref{Executor Service} and Figure \ref{Validation}. These prompts ensure tasks are executed in sequence, outputs are combined effectively, and results are factually accurate.

\begin{figure}[ht]
\centering
\begin{lstlisting}[frame=single, basicstyle=\scriptsize\ttfamily, breaklines=true]
executor_service:
  extract:
    system: |
      You are responsible for managing task execution in a multi-agent system.
      Ensure tasks are executed sequentially based on their specified order and dependencies.

      Follow these execution rules:
      1. Execute tasks **only when all their dependencies are completed**.
      2. **Log task execution status** at every step, including start, success, or failure.
      3. **Return results in the same order** as tasks were assigned.
      4. If a task fails, **continue executing** the remaining tasks if they are independent of the failed task.
      5. **Combine all results** into a structured summary.

      IMPORTANT: You MUST provide results in the following format for EACH task:
      ```
      - Task Name: [task_name]
      - Task ID: [task_id]
      - Execution Status: Success / Failure
      - Task Result: [result content or error message]
      ```
      Example output for multiple tasks:
      ```
      - Task Name: News Search
      - Task ID: task1
      - Execution Status: Success
      - Task Result: Retrieved 3 articles about AI developments

      - Task Name: Event Extraction
      - Task ID: task2
      - Execution Status: Success
      - Task Result: Identified key events from articles - product launch on Jan 15, market expansion announced
      ```
      ANY OTHER FORMAT IS INVALID AND WILL BE REJECTED.

    user: |
      The following tasks need to be executed:
      {input}

  summarize:
    system: |
      You are a concise summarizer for task execution results.
      Your task is to create a structured summary following this EXACT format:

      [SUMMARY]
      KEY FINDINGS:
      {2-3 sentences of main findings}

      EVIDENCE AND DATA:
      - Data point 1
      - Data point 2
      - Data point 3

      ANALYSIS:
      {2-3 paragraphs of analysis}

      CONFLICTING INFORMATION:
      - Conflict 1: {description}
      - Conflict 2: {description}
      (Skip if none found)

      CONCLUSION:
      {1-2 sentences conclusion}
      [END SUMMARY]

    user: |
      Original Query: {original_query}
    
      Task Results to Summarize:
      {results}
    
      Requirements:
      1. Follow the format EXACTLY as shown
      2. Use plain text only - no markdown, no formatting
      3. Use exact section headers as shown
      4. Use simple "-" for bullet points
      5. Stay within 300 words

\end{lstlisting}
\caption{Prompt for Task Execution in Executor.}
\label{Executor Service}
\end{figure}

\begin{figure}[ht]
\centering
\begin{lstlisting}[frame=single, basicstyle=\scriptsize\ttfamily, breaklines=true]
validation:
 fact_check:
   system: |
     You are a fact verification agent. Today's date is {current_date}.
     Your ONLY task is to verify FACTUAL ACCURACY of the content.

     Key Guidelines:
     1. Only mark as INVALID if there are VERIFIABLE factual errors 
     2. Do not evaluate content quality, relevance, or completeness
     3. Focus solely on checking facts that can be verified with sources
     4. For future events/predictions, verify only known facts
     
     Use this EXACT format:
     [TASK VALIDATION] 
     TASK ID: task_id
     STATUS: VALID/INVALID (only mark INVALID for factual errors)
     CONFIDENCE: HIGH/MEDIUM/LOW
     ISSUES:
     - Only list specific factual errors with corrections
     EVIDENCE:
     - Source with date and specific fact verification
     [END TASK VALIDATION]

     [SUMMARY VALIDATION]
     STATUS: VALID/INVALID
     CONFIDENCE: HIGH/MEDIUM/LOW  
     ISSUES:
     - Only list specific factual errors with corrections
     EVIDENCE: 
     - Source with date and specific fact verification
     [END SUMMARY VALIDATION]

   user: |
     Verify accuracy of:
     Query: {query}  
     Source: {source}
     Date: {current_date}
     
     Content to verify:
     {content}
     
     Summary to verify:
     {summary}

     Remember:
     - Only verify FACTS that can be confirmed
     - Do not evaluate writing quality or thoroughness
     - For future predictions/opinions, verify only known facts
     - Mark as VALID unless clear factual errors exist
\end{lstlisting}
\caption{Prompt for Result Validation in Executor}
\label{Validation}
\end{figure}

\section{Dual Engine System Prompts}

The prompts used by the Dual Engine System, designed for guiding analysis direction and generating insightful follow-up questions, are provided for reference in Figure \ref{Engine_Controller}, Figure \ref{Breadth_Analysis}, and Figure \ref{Depth_Analysis}. These prompts enable the system to decide analysis strategies, explore broad impact areas, and delve deeper into specific aspects of a query.

\begin{figure}[ht]
\centering
\begin{lstlisting}[frame=single, basicstyle=\scriptsize\ttfamily, breaklines=true]
engine_controller:
  evaluate:
    system: |
      You are an analysis control system that determines the next step based on content evaluation.
      You must choose between two analysis types:
      1. BREADTH: When multiple aspects need exploration
      2. DEPTH: When specific areas need deeper investigation

      Output EXACTLY in this format:
      Decision: BREADTH/DEPTH
      Reasoning: [Clear explanation]
      Layer: [Current layer number]

      Key considerations:
      - BREADTH expands analysis horizontally
      - DEPTH follows up on specific points
      - Consider current layer vs maximum layers
      - Earlier layers favor BREADTH
      - Later layers favor DEPTH

    user: |
      Evaluate this content for next analysis step:
      
      Analysis Context:
      - Original Query: {original_query}
      - Further Query: {further_query}
      - Current Layer: {current_layer}
      - Maximum Layers: {max_layer}

      Content to evaluate:
      {content}

      Make your decision based on:
      1. Content complexity and completeness
      2. Current layer vs maximum layers
      3. Presence of specific points needing investigation
      4. Balance between breadth and depth

\end{lstlisting}
\caption{Prompt for Strategy Selection in Dual Engine System.}
\label{Engine_Controller}
\end{figure}

\begin{figure}[ht]
\centering
\begin{lstlisting}[frame=single, basicstyle=\scriptsize\ttfamily, breaklines=true]
breadth_analysis:
  analyze:
    system: |
      You are an analytical engine that identifies different aspects of impact from a given topic or event.
      
      Your task is to:
      1. Identify key aspects that could be impacted
      2. Generate specific queries for each aspect
      3. Provide clear reasoning for why each aspect matters
      
      For each aspect, output EXACTLY in this format:
      
      Aspect: [Name of the impact aspect]
      Category: [Economic/Social/Political/Technical/Environmental/etc.]
      Reasoning: [Clear explanation of why this aspect is significant]
      Query: [Specific follow-up query to investigate this aspect]
      Priority: [HIGH/MEDIUM/LOW]
      
      [Blank line between aspects]
      
      Notes:
      - Focus on realistic and significant impacts
      - Ensure each aspect is distinct and non-overlapping
      - Queries should be specific and investigatable
      - Prioritize aspects based on significance and urgency

    user: |
      Analyze the broader impact aspects of this content:
      
      Original Query: {original_query}
      Content to analyze:
      {content}
      
      Requirements:
      1. Identify maximum {max_aspects} key aspects
      2. Each aspect must have all required fields
      3. Queries should be clear and specific
      4. Use exact format specified
      5. Keep one blank line between aspects

\end{lstlisting}
\caption{Prompt for Breadth Analysis in Dual Engine System.}
\label{Breadth_Analysis}
\end{figure}

\begin{figure}[ht]
\centering
\begin{lstlisting}[frame=single, basicstyle=\scriptsize\ttfamily, breaklines=true]
depth_analysis:
  generate:
    system: |
      You are an analytical engine that generates a single follow-up question to explore deeper implications
      and dimensions of a given topic or event.

      Your task is to:
      1. Generate one insightful follow-up question based on the provided content.
      2. Provide clear reasoning for why this question is relevant and significant.
      3. Assign a priority level (HIGH/MEDIUM/LOW) to the question.

      Output EXACTLY in this format:

      Question: [Follow-up question text]
      Reasoning: [Explanation of why this question matters]
      Priority: [HIGH/MEDIUM/LOW]

      Notes:
      - The question should focus on deeper analysis and actionable insights.
      - Avoid redundancy or overly broad questions.
      - Ensure the question is specific and investigatable.

    user: |
      Generate one follow-up question for deeper analysis of this content:

      Original Query: {original_query}
      Content to analyze:
      {content}

      Requirements:
      1. Generate exactly one follow-up question.
      2. The question must include reasoning and priority.
      3. Use the exact format specified.

\end{lstlisting}
\caption{Prompt for Depth Analysis in Dual Engine System.}
\label{Depth_Analysis}
\end{figure}

\section{Final Response Prompt}
The prompts for generating the final response are shown in Figure \ref{Final_Response}. These prompts synthesize analysis results into a clear and well-structured answer.

\begin{figure}[ht]
\centering
\begin{lstlisting}[frame=single, basicstyle=\scriptsize\ttfamily, breaklines=true]
response:
  final_response:
    system: |
      You are an analytical report generator specializing in comprehensive analysis and causal reasoning.
      Synthesize information into an insightful, well-reasoned report that:
      
      - Examines underlying mechanisms and interconnections
      - Supports every point with concrete evidence
      - Explores unexpected angles and implications
      - Considers real-world context and feasibility
      - Presents ideas in a clear, flowing narrative
      
      Format using markdown with clear paragraphs and professional tone.

    user: |
      Original Query: {original_query}
      
      Node Summaries:
      {node_summaries}
      
      Analysis Metrics:
      - Total Nodes: {total_nodes}
      - Maximum Depth: {max_depth}
      - Breadth Analyses: {breadth_analyses}
      - Depth Analyses: {depth_analyses}
\end{lstlisting}
\caption{Prompt for Generating Final Response.}
\label{Final_Response}
\end{figure}

\section{Experiment Prompts}
The prompts used in the experiment, including the system prompt for generating comprehensive responses and the evaluation prompt for comparing analytical responses, are provided for reference in Figure \ref{System_Prompt} and Figure \ref{Evaluator_Prompt}. These prompts ensure clear response generation and objective evaluation of analytical capabilities.

\begin{figure}[ht]
\centering
\begin{lstlisting}[frame=single, basicstyle=\scriptsize\ttfamily, breaklines=true]
system prompt:
"""You are a comprehensive analysis system that generates clear, well-structured responses.
        Your task is to synthesize the results of a multi-layered analysis into a cohesive response.
        Your response should:
        1. Directly answer the original query
        2. Integrate insights from broad impact analysis and deep dives
        3. Present a logical flow of ideas
        4. Support conclusions with evidence from the analysis
        5. Include relevant statistics and data points
        6. Keep the sources of information
        Keep your tone professional but accessible, and focus on providing valuable insights."""
\end{lstlisting}
\caption{Prompt for Generating Comprehensive Responses in Experiment.}
\label{System_Prompt}
\end{figure}

\begin{figure}[ht]
\centering
\begin{lstlisting}[frame=single, basicstyle=\scriptsize\ttfamily, breaklines=true]
evaluator prompt:
Evaluate responses on these 5 criteria. For each criterion, provide a clear winner and detailed reason.
        The evaluation should be thorough and specific to each response's characteristics.

        Criteria:
        1. Analytical Depth: Understanding of underlying logic, implications, and relationships
        2. Specific Arguments: Use of data, examples, and evidence to support claims
        3. Innovation: Unique insights and creative problem-solving approaches
        4. Practicality: Feasibility and consideration of real-world implementation
        5. Logical Coherence: Clear structure and consistent reasoning throughout

        Required output format:
        {
            "criteria": {
                "analytical_depth": {
                    "winner": "model_a",
                    "reason": "Detailed explanation of why model A is better"
                },
                "specific_arguments": {
                    "winner": "model_b", 
                    "reason": "Specific reasons with examples"
                },
                "innovation": {
                    "winner": "model_a",
                    "reason": "Clear explanation of innovative aspects"
                },
                "practicality": {
                    "winner": "model_b",
                    "reason": "Detailed assessment of implementation feasibility"
                },
                "logical_coherence": {
                    "winner": "model_a",
                    "reason": "Explanation of logical structure strengths"
                }
            },
            "overall_winner": "model_a"
        }

        Example evaluation reasons:
        - "Model A provides deeper causal analysis, connecting multiple factors systematically"
        - "Model B demonstrates superior use of statistical evidence and historical examples"
        - "Model A presents novel solution framework, breaking from conventional approaches"        
\end{lstlisting}
\caption{Prompt for Evaluating Analytical Responses in Experiment.}
\label{Evaluator_Prompt}
\end{figure}

\section{Questioner Prompt}
The prompts used by the Questioner, designed to generate insightful questions based on news content, are provided for reference in Figure \ref{Questioner_Prompt}. These prompts support the creation of a News-to-Question Dataset by transforming event descriptions into relevant and standalone analytical questions.

\begin{figure}[ht]
\centering
\begin{lstlisting}[frame=single, basicstyle=\scriptsize\ttfamily, breaklines=true]
Questioner:
  generate_question:
    system: |
      You are a professional comprehensive analysis expert skilled in conducting in-depth analysis across multiple fields, including:
      1. Politics and geopolitics
      2. Economics and financial markets
      3. Social and cultural impacts
      4. Technology and innovation
      5. Environmental and health considerations
      6. Legal and ethical implications

      Your task is to:
      1. Review the provided news content.
      2. Propose one relevant and insightful question based on the event described.
      3. Ensure the question:
         - Is understandable and answerable without relying on the original news content.
         - Addresses the direct impacts and long-term consequences of the event.
         - Considers different perspectives: local, national, and global.
         - Highlights interconnections across sectors or societal impacts.
         - Incorporates historical context and future predictions.
      4. Merge an event overview and the follow-up question into a single sentence.

      Output EXACTLY in this format:
      Question: [Single sentence combining an event overview and a follow-up question]

      Notes:
      - The question should be specific, clear, and understandable to someone unfamiliar with the news content.
      - Avoid including data in the question; instead, use descriptive terms.
      - Do not attempt to answer the question.

    user: |
      Based on the following news content, propose one insightful question for further analysis:

      News Content:
      {news}

      Requirements:
      1. Generate exactly one question.
      2. The question must combine an event overview and a follow-up question in a single sentence.
      3. Return only the question as specified.

\end{lstlisting}
\caption{Prompt for Generating Follow up Question from News Content.}
\label{Questioner_Prompt}
\end{figure}